%% file: PhysWM_RSS.tex
\documentclass[conference]{IEEEtran}
\usepackage{times}

\usepackage[numbers]{natbib}
\usepackage{multicol}
\usepackage[bookmarks=true]{hyperref}
\usepackage{xcolor}         % colors
\usepackage{graphicx}
\usepackage{multirow}
\usepackage{booktabs}
\usepackage{caption}
\usepackage{amssymb}
\usepackage{amsmath}
\usepackage{bm}
\usepackage{subfigure}
\usepackage{overpic}
\usepackage{rotating}

\begin{document}

\definecolor{green}{rgb}{0, 0.5, 0}
\definecolor{orange}{rgb}{1.0, 0.6, 0.2}
\definecolor{red}{rgb}{1.0, 0.0, 0.0}
\definecolor{blue}{rgb}{0.0, 0.0, 1.0}
\definecolor{teal}{rgb}{0.0, 0.4, 0.4}
\definecolor{purple}{rgb}{0.65,0,0.65}
\definecolor{saffron}{rgb}{0.95,0.75,0.2}
\definecolor{turquoise}{rgb}{0.0,0.5,0.5}
\definecolor{black}{rgb}{0,0,0}

\newcommand{\kx}[1]{{\color{red}#1}}
\newcommand{\zh}[1]{{\color{orange}#1}}
\newcommand{\rh}[1]{{\color{blue}#1}}
\newcommand{\sure}[1]{{\color{green}#1}}
\newcommand{\unsure}[1]{{\color{purple}#1}}

\newcommand{\method}{PIN-WM}

% \begingroup
\title{
\begingroup
\setlength{\spaceskip}{0.25em}
PIN-WM: Learning Physics-INformed World Models \\ for  Non-Prehensile Manipulation
\endgroup
}
% \endgroup

\author{\authorblockN{Wenxuan Li$^{1,*}$\quad Hang Zhao$^{2,*}$\quad Zhiyuan Yu$^2$\quad Yu Du$^1$\quad Qin Zou$^{2, 4}$\quad Ruizhen Hu$^{3,\dag}$\quad Kai Xu$^{1,\dag}$
}
\authorblockA{
   $^1$National University of Defense Technology \quad  $^2$Wuhan University \\ $^3$Shenzhen University  \quad $^4$Guangdong Laboratory of Artificial Intelligence and Digital Economy \\
   $^*$Equal contributions \quad $^\dag$Corresponding author
   \\
   Project page:  \href{https://pinwm.github.io}{\texttt{https://pinwm.github.io}}
}
}

% avoiding spaces at the end of the author lines is not a problem with
% conference papers because we don't use \thanks or \IEEEmembership

%Homer Simpson\authorrefmark{2},
%Montgomery Scott\authorrefmark{3} and
%Eldon Tyrell\authorrefmark{4}}
%Telephone: (800) 555--1212, Fax: (888) 555--1212}
%\authorblockA{\authorrefmark{4}Tyrell Inc., 123 Replicant Street, Los Angeles, California 90210--4321}}

\maketitle

\input{abstract.tex}

\IEEEpeerreviewmaketitle

\input{introduction.tex}

\input{related_work.tex}

\input{method.tex}

\input{result.tex}

\input{conclusion.tex}

%\clearpage

%% Use plainnat to work nicely with natbib. 

\bibliographystyle{plainnat}
\bibliography{references}

\input{appendix.tex}

\end{document}

%% file: abstract.tex
\begin{abstract}
    While non-prehensile manipulation (e.g., controlled pushing/poking) constitutes a foundational robotic skill, its learning remains challenging due to the high sensitivity to complex physical interactions involving friction and restitution. To achieve robust policy learning and generalization, we opt to learn a world model of the 3D rigid body dynamics involved in non-prehensile manipulations and use it for model-based reinforcement learning.
    We propose \method{}, a Physics-INformed World Model that enables efficient end-to-end identification of a 3D rigid body dynamical system from visual observations. Adopting differentiable physics simulation, \method{} can be learned with only few-shot and task-agnostic physical interaction trajectories.
    Further, \method{} is learned with observational loss induced by 
    Gaussian Splatting without needing state estimation. To bridge Sim2Real gaps, we turn the learned \method{} into a group of Digital Cousins via physics-aware randomizations which perturb physics and rendering parameters to generate diverse and meaningful variations of the \method{}.
    Extensive evaluations on both simulation and real-world tests demonstrate that PIN-WM, enhanced with physics-aware digital cousins, facilitates learning robust non-prehensile manipulation skills with Sim2Real transfer, surpassing the Real2Sim2Real state-of-the-arts.
\end{abstract}

%% file: introduction.tex
\section{Introduction}
\label{sec:introduction}

Non-prehensile robotic manipulation~\cite{mason1986mechanics, gondokaryono2023learning, zhou2023hacman}, which involves moving an object by pushing or poking, finds extensive applications in many real-world scenarios where grasping is infeasible due to the weight, size, shape, or fragility of the object, among others.
Robotic push can be implemented with simpler end effectors, making systems more cost-effective and easier to deploy in certain environments~\cite{ebel2022finding, SongB20}.
However, significant challenges arise from the difficulty of fully dictating the motion and pose of the object being pushed.
The complex underlying dynamics, caused by factors such as friction, restitution, and inertia, make motion prediction difficult and complicate motion planning and control.

\begin{figure}[htp]
    \centering
    \includegraphics[width=\linewidth]{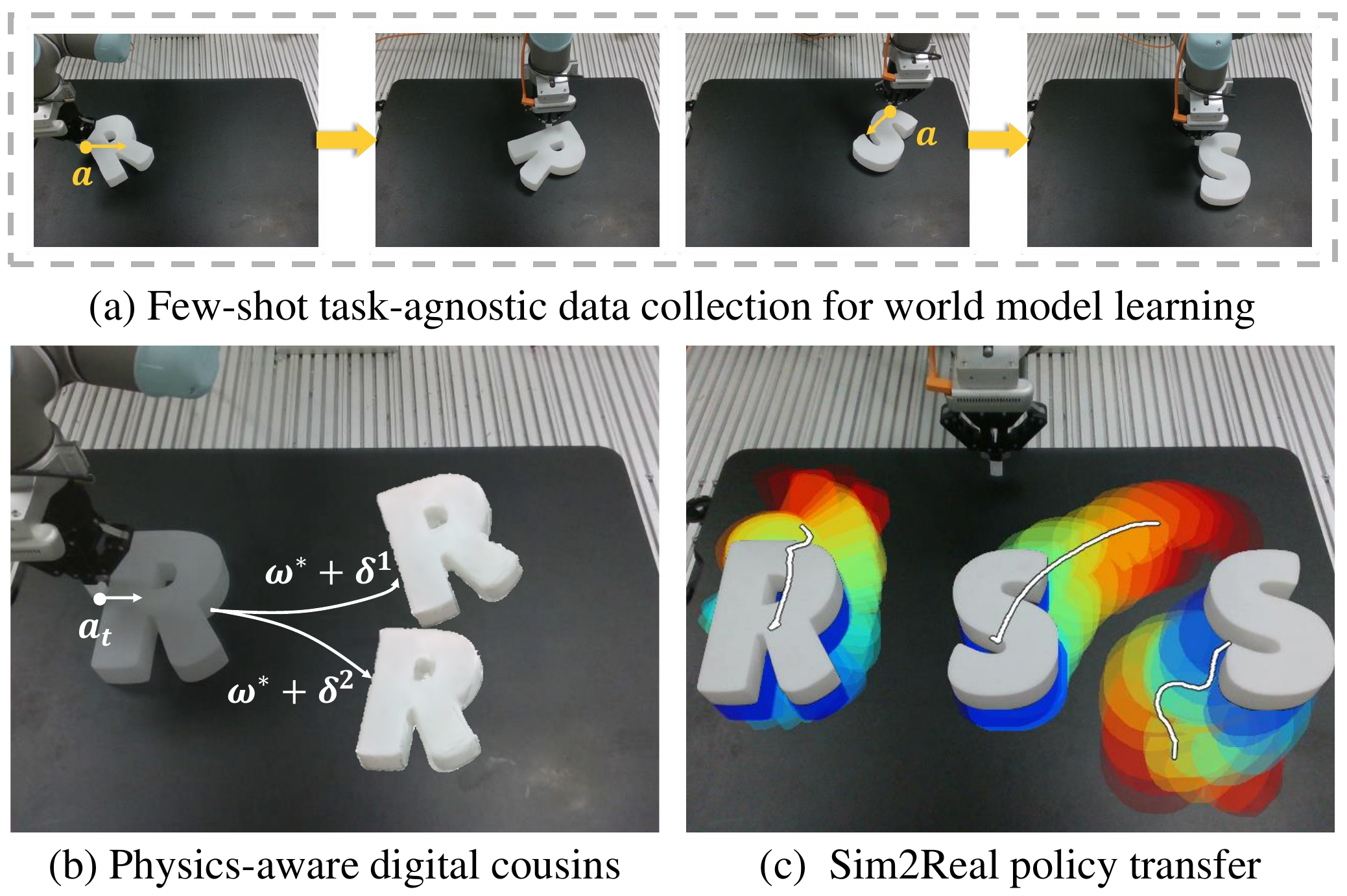}
    \caption{
    PIN-WM is learned from few-shot and task-agnostic physical interaction trajectories (random pushes of the blocks in this example), through end-to-end differentiable identification of 3D physics parameters essential to the push operation (a). The learned PIN-WM is then turned into a group of digital cousins via physics-aware perturbations (b). The resulting world models are then used to learn the task-specific policies with Sim2Real transferability (c).
    }
    \label{fig:teaser}
\end{figure}

Some studies tackle non-prehensile manipulation with imitation learning~\cite{chi2023diffusion, young2021visual}, where the reliance on expensive expert demonstrations limits their scalability.  Others explore deep reinforcement learning (DRL)~\cite{sutton2018reinforcement}, leveraging trial-and-error in simulations to learn policies~\cite{kalashnikov2018scalable, yuan2018rearrangement}. However, the discrepancy between simulation and reality hinders the transfer of the learned policy to real-world environments~\cite{clavera2017policy, li2024robogsim}.  A promising alternative is to learn world models~\cite{ha2018world} of the environment dynamics in a data-driven manner, which can be used for predictive control or employed in model-based RL for better data-efficiency and Sim2Real generality~\cite{HafnerLB020, HansenSW22}.
However, purely data-driven world models rely heavily on the quantity and quality of training data and struggle to generalize to out-of-distribution (OOD) scenarios~\cite{yu2020mopo}.

It is well-recognized that incorporating structured priors into learning algorithms improves generalization with limited training examples~\cite{raissi2019physics}.
A number of studies~\cite{memmel2024asid,  baumeister2024incremental, SongB20} have sought to integrate principles of physics into the development of world models. In doing so, two critical aspects require particular consideration. \emph{The first is the differentiability of the physics parameter identification process}.
ASID~\cite{memmel2024asid} identifies physics parameters for an established simulator using gradient-free optimization~\cite{rubinstein2004cross}.
\citet{baumeister2024incremental} adopt a similar method to learn dynamics for model predictive control (MPC)~\cite{williams2017model}.
The absence of gradient feedback renders these methodologies critically dependent on data quality. However, collecting high-quality real-world trajectories itself is a challenging task~\cite{memmel2024asid}. 
\citet{SongB20} employ a differentiable 2D physics simulator for learning planar sliding dynamics. However, 2D physics is insufficient to capture complex motions such as flipping an object through poking.
\emph{The second consideration lies in the necessity of state estimation in the optimization of world models.}
While most existing methods involve state estimation with additional modules~\cite{ memmel2024asid,baumeister2024incremental, SongB20}, the recent advances in differentiable rendering~\cite{kerbl20233d, wu2024recent, mu2023neural} make it possible to optimize against an observational loss directly, thus saving the effort on state estimation.

We introduce  \textbf{\method}, a \textbf{P}hysics-\textbf{IN}formed \textbf{W}orld \textbf{M}odel that allows end-to-end identification of a 3D rigid body dynamical system from visual observations.
\emph{First}, \method{} is a differentiable approach to the identification of 3D physics parameters, requiring only few-shot and task-agnostic physical interaction trajectories.
Our method systematically identifies critical dynamics parameters essential to non-prehensile manipulation, encompassing inertial properties, frictional coefficients, and restitution characteristics~\cite{SongB20, memmel2024asid, ferrandis2024learning}.
\emph{Second}, \method{} learns physics parameters by optimizing the rendering loss~\cite{nerf} induced by the 3D Gaussian Splatting~\cite{2DGS} scene representation, facilitating direct learning from RGB images without additional state estimation modules. Consequently, the learned world model, with identified physics and rendering properties, can be readily applied to train vision-based control policies of non-prehensile manipulation using RL.

The learned \method{}, representing a digital twin~\cite{grieves2017digital} of the real-world rigid body system,
may still exhibit discrepancies against reality due to the inaccurate and partial observations~\cite{kinoshita2000robotic}.
To bridge the Sim2Real gap, we turn the identified digital twin into plenty of digital cousins~\cite{Digital_Cousins} through physics-aware perturbations which perturb the physics and rendering parameters around the identified values as means. Such purposeful randomization creates a group of \emph{physics-aware digital cousins} obeying physics laws while introducing adequate varieties accounting for the unmodeled discrepancies. 
The resulting world model group allows learning robust non-prehensile manipulation policies with Sim2Real transferability.

Through extensive evaluation across diverse task scenarios, we demonstrate that \method{} is fast-to-learn and accurate, making it useful in learning robust non-prehensile manipulation skills with strong Sim2Real transfer. The overall performance surpasses the recent Real2Sim2Real state-of-the-arts~\cite{SongB20, memmel2024asid, li2024robogsim} significantly.
Our real-world experiments further showcase that \method{} facilitates Sim2Real policy transfer without real-world fine-tuning and achieves high success rates of $75\%$ and $65\%$ in the \textit{Push} and \textit{Flip} tasks, respectively. 
Our contributions include:
\begin{itemize}
    \item We propose \method{} for accurate and efficient identification of world models of 3D rigid body dynamical systems from visual observations in an end-to-end fashion.
    \item We turn the identified digital twin into a group of physics-aware digital cousins through perturbing the physics and rendering parameters around the identified mean values, to support learning non-prehensile manipulation skills with robust Sim2Real transfer.
    \item We conduct real robot implementation to demonstrate that our approach enables learning control policies with minimal task-agnostic interaction data and attains high performance Real2Sim2Real without real-world fine-tuning.
\end{itemize}

%% file: related_work.tex
\section{Related Work}
\label{sec:related_work}

\subsection{Non-Prehensile Manipulation}

Non-prehensile manipulation~\cite{mason1986mechanics, gondokaryono2023learning, zhou2023hacman} refers to controlling objects without fully grasping them.  
While offering flexibility, its motions are highly sensitive to contact configurations~\cite{yu2016more}, requiring accurate dynamic descriptions and control policies.
\citet{mason1986mechanics} presents a theoretical framework for pushing mechanics and derive the planning by predicting the rotation and translation of an object pushed by a point contact.
\citet{akella1998posing} use a theoretical guaranteed linear programming algorithm
to solve pose transitions and generate open-loop push plans without sensing requirements. 
\citet{dogar2011framework} adopt an action library and combinatorial search inspired by human strategies. The library can rearrange cluttered environments using actions such as pushing, sliding, and sweeping.
\citet{zhou2019pushing}  model pushing mechanisms use sticking contact and an ellipsoid approximation of the limit surface, enabling path planning by transforming sticking contact constraints into curvature constraints.
These methods, however, rely on simplified assumptions, either known physics parameters or idealized physical models, which are often violated in practice~\cite{dynamicsDM}. 

Deep learning methods have recently been applied to train non-prehensile policies. 
Some studies focus on imitation learning~\cite{hussein2017imitation}, which mimics expert behavior for specific tasks.
\citet{young2021visual} emphasize the importance of diverse demonstration data for generalizing non-prehensile manipulation tasks, introducing an efficient visual imitation interface that achieves high success rates in real-world robotic pushing.
\citet{chi2023diffusion} utilize diffusion models' multimodal action distribution capabilities~\cite{ddpm} to imitate pushing T-shaped objects, demonstrating impressive robustness. 
While effective, imitation learning relies heavily on the quantity of real-world data.
Otherwise, it is prone to state-action distribution shifts during sequential decision-making~\cite{belkhale2024data}, which is a critical issue in non-prehensile tasks requiring precise contact point selection and control~\cite{zhou2023hacman}.
\citet{hudata} conclude that imitation generalization follows a scaling law~\cite{kaplan2020scaling} with the number of environments and objects, recommending 50 demonstrations per environment-object pair. 
Such data requirements can be costly and prohibitive for scalability.
Alternatively, deep reinforcement learning (DRL) can learn policies through trial and error in simulated environments~\cite{kalashnikov2018scalable, yuan2018rearrangement}. However, the large gap between simulation and reality poses significant challenges for transferring these policies to the real world~\cite{clavera2017policy, li2024robogsim}.
Building an interactive model that accurately captures real-world physical laws is crucial for learning feasible non-prehensile manipulation policies in real world.

\subsection{World Models for Policy Learning}

World models~\cite{ha2018world}, which learn the environment dynamics in a data-driven manner, provide interactive environments for effective policy training~\cite{HafnerLB020, HansenSW22}.
\citet{HafnerLB020} propose Dreamer, a world model that learns a compact latent representation of the environment dynamics.
The following work~\cite{WuEHAG22} applies Dreamer to robotic manipulation tasks, demonstrating fast policy learning on physical robots. 
DINO-WM~\cite{zhou2024dino} leverages spatial patch features pre-trained with DINOv2  to learn a world model and achieve task-agnostic behavior planning by treating goal features as prediction targets.
TD-MPC~\cite{HansenSW22, 00010024} uses a task-oriented latent dynamics model for local trajectory optimization and a learned terminal value function for long-term return estimation, achieving superiority on image-based control.
Building on the success of learning from large-scale datasets~\cite{brown2020language, radford2021learning}, \citet{MendoncaBP23}  leverage internet-scale video data to learn a human-centric action space grounded world model.
However, purely data-driven world models rely heavily on the quantity and quality of training data and struggle to generalize to out-of-distribution (OOD) scenarios~\cite{yu2020mopo, rafailov2021offline}. This lowers the robustness of the learned policies transferred to the real world.

Incorporating structured priors into learning algorithms is known to improve generalization with limited training data~\cite{raissi2019physics, caoneuma}.
Recent advances in differentiable physics have opened up new possibilities for incorporating physical knowledge into world models.
\citet{LutterRP19} introduce a deep network framework based on Lagrangian mechanics, efficiently learning equations of motion while ensuring physical plausibility. 
\citet{heiden2021neuralsim} augment a differentiable rigid-body physics engine with neural networks to capture nonlinear relationships between dynamic quantities.
Other works \cite{2DLCP} demonstrate analytical backpropagation through a physical simulator defined via a linear complementarity problem.
$\nabla$Sim~\cite{MurthyMGVPWCPXE21} combine differentiable physics~\cite{2DLCP, 3DLCP} and rendering~\cite{chen2024neural,jing2023state,jing2024frnerf, yang2023jnerf} to jointly model scene dynamics and image formation, enabling backpropagation from video pixels to physical attributes. 
This approach was soon followed by improvements with advanced rendering techniques~\cite{LiQCJLJG23,caoneuma}, or enhanced physics engines~\cite{kandukuri2024physics}.

Despite those advances, only a few studies~\cite{memmel2024asid, baumeister2024incremental, SongB20} incorporate physical property estimation into world models for non-prehensile manipulation, relying on gradient-free optimization or simplified physical models that fail to effectively handle complex interactions. 
Gradient-free methods rely on high-quality trajectories for system identification; lacking such data, they are prone to local optima, as demonstrated by ASID using CEM~\cite{memmel2024asid}. 
Simplified physics models, such as the 2D physics engine adopted by \citet{SongB20}, inherently struggle to capture the full 3D dynamics of real-world interactions, leading to inaccurate predictions.
In contrast, \method{} enables end-to-end identification of 3D rigid-body dynamics from visual observations using few-shot, task-agnostic interaction data, which facilitates the training of vision-based manipulation policies with RL.
PIN-WM aligns with the original,
narrow-scope definition of a world model~\cite{ha2018world}: a dynamics model tailored to a specific environment for precise model-based control. This contrasts with general-purpose world foundation models like Cosmos~\cite{agarwal2025cosmos}.

\subsection{Domain Randomization}   

Domain Randomization trains a single policy across a range of environment parameters to achieve robust performance during testing. 
\citet{dynamicsDM} enhance policy adaptability to varying environmental dynamics by randomizing the environment's dynamic parameters. 
\citet{miki2022learning} train legged robots in diverse simulated physical environments. During testing, the robots first probe the terrain through physical contact, then preemptively plan and adapt their gait, resulting in high robustness and speed.
\citet{visionDM} introduce randomized rendering, e.g., textures, lighting, and backgrounds,  in simulated environments to enhance real-world visual detection.
\citet{yue2019domain} randomize synthetic images using real image styles from auxiliary datasets to learn domain-invariant representations. 
\citet{Digital_Cousins} introduce an automated pipeline to transform real-world scenes into diverse, interactive digital cousin environments, demonstrating significantly higher transfer success rates compared to digital twins~\cite{grieves2017digital}.

While these randomization methods provide a simple approach for efficiently transferring simulation-trained policies to the real world, their uniform sampling of environment parameters lacks proper constraints. This results in a generated space far larger than the real-world space, increasing learning burdens~\cite{mehta2020active} and often producing conservative policies with degraded performance~\cite{evans2022context}. 
In contrast, we perturb the physics and rendering parameters around the identified values as means.
Such purposeful randomization creates a group of \emph{physics-aware digital cousins} obeying physics laws while introducing adequate varieties accounting for the unmodeled discrepancies. 
The developed world model facilitates the learning of robust non-prehensile manipulation policies that transfer effectively from simulation to real-world environments.

%% file: method.tex
\section{Method}
\label{sec:method}

\subsection{Overall Framework}

\begin{figure*}[htp]
    \centering
    \includegraphics[width=\linewidth]{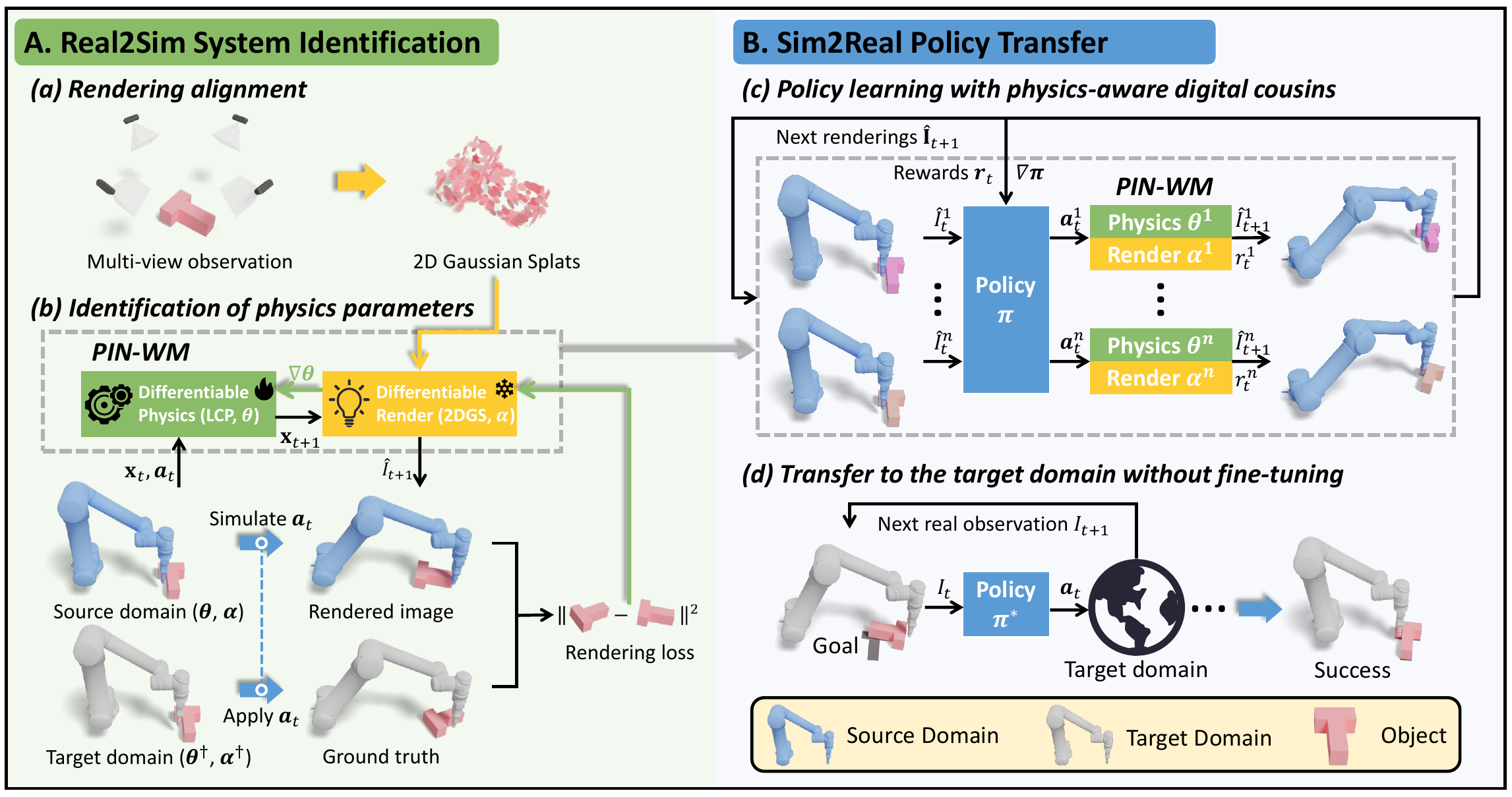}
    \caption{
        Our Real2Sim2Real framework for learning non-prehensile manipulation policies.
        (a) The robot in the target domain moves around the object, capturing multi-view observations to estimate the rendering parameters $\bm{\alpha}$ of 2D Gaussian Splats. (b) Once optimized, $\bm{\alpha}$ is frozen. Both source and target domains apply the same task-agnostic physical interactions $\mathbf{a}_t$. In the source domain,  dynamics are computed via LCP  with physical parameters $\bm{\theta}$ to update the rendering. $\bm{\theta}$ is then optimized with the rendering loss between two domains.
        (c) The identified world model is then used for policy learning. Physics-aware perturbations are introduced to $\bm{\alpha}$ and $\bm{\theta}$ to mitigate the remained discrepancies from inaccurate observations.
        (d) This ensemble of perturbed world models enhances the Sim2Real transferability of learned policies. 
    }
    \label{fig:pipeline}
    \vspace{-14pt}
\end{figure*}

We develop real-world non-prehensile manipulation skills through a two-stage pipeline: Real2Sim system identification via our physics-informed world model, 
 and Sim2Real policy transfer enhanced by physics-aware digital cousins. We provide an overview of our framework in Figure~\ref{fig:pipeline}.

\paragraph*{\textbf{Real2Sim System Identification}}
The Real2Sim stage constructs our physics-informed world model, which identifies the physics parameters of the target domain from visual observations.  
A world model~\cite{HafnerLB020}  predicts the next system observation $\mathbf{o}_{t+1}$ based on the current observation $\mathbf{o}_{t}$ and actions $\mathbf{a}_t$:
\begin{equation}
    \mathbf{o}_{t+1} = \mathcal{W}(\mathbf{o}_t, \mathbf{a}_t, \bm{\omega}),
\end{equation}
where $t$ denotes the time step and $\bm{\omega}$ represents the learnable parameters.
A comprehensive physical world for robot interaction should account for visual observations, physics, and geometry~\cite{abouphysically}, and we follow the convention in previous non-prehensile manipulation works~\cite{SongB20,zhou2023hacman} that the geometry is assumed to be known. 
Therefore, our \method{}   $\mathcal{W} = \mathcal{I} \circ g$ 
focuses on learning visual observations and physics of the target domain, where $\mathcal{I}$ is the differentiable rendering function 
and $g$ is the differentiable physics function. In more detail, the world model can be rephrased as:
\begin{equation}
\label{eq:wm}
    I_{t+1} = \mathcal{I}(g(\mathbf{x}_t, \mathbf{a}_t, \bm{\theta}), \bm{\alpha}), 
\end{equation}
where $g$, parameterized by $\bm{\theta}$, predicts the next state $\mathbf{x}_{t+1}$ from current state $\mathbf{x}_{t}$ and action $\mathbf{a}_t$, and $\mathcal{I}$, parameterized by $\bm{\alpha}$, generates the image $I_{t+1}$ corresponding to $\mathbf{x}_{t+1}$.
Hence, $\bm{\omega} = \{\bm{\alpha}, \bm{\theta} \}$ forms all learnable parameters for $\mathcal{W}$. The goal of system identification is to optimize $\bm{\omega}$ so that the generated images 
resemble those observed in the target domain.

\paragraph*{\textbf{Sim2Real Policy Transfer}}
After system identification, we obtain the world model $\mathcal{W}$ as an interactive simulation environment. 
We can learn non-prehensile manipulation skills through reinforcement learning, where the learned policy is expected to achieve Sim2Real transfer without real-world fine-tuning. 
However, the identified world model may deviate from the real world due to inaccurate and partial observations~\cite{kinoshita2000robotic}.
We enhance policy transfer performance by introducing 
\emph{physics-aware digital cousins} (PADC).
PADC perturbs the identified system to generate meaningful training variations, which share similar physics and rendering properties while introducing distinctions to model unobserved discrepancies.
This approach improves policy transferability and reduces the learning burden. 
The learned policy is then directly deployed in the target domain for manipulation tasks.

\subsection{Physics-INformed World Model}

In this section, we provide a detailed description of learning PIN-WM $\mathcal{W} = \mathcal{I} \circ g$.
To fully characterize the dynamics $g$ of our system, we adopt rigid body simulation~\cite{3DLCP} to formulate the dynamics of scene components that satisfy the momentum conservation. Therefore, we include the target object, the end-effector, and the floor in our environment state representation $\mathbf{x}_t = \{\mathbf{p}_t, \mathbf{q}_t, \bm{\xi}_t \}$, where $\mathbf{p}_t$, $\mathbf{q}_t$, and $\bm{\xi}_t$ represents their positions, orientations, and twist velocities, respectively.

We account for joint, contact, and friction constraints for rigid body simulation, and include the physical properties of those scene components that are most concerned by non-prehensile manipulation tasks~\cite{SongB20, memmel2024asid} into our physical parameters $\bm{\theta} = \{\bm{\theta}^\mathbf{M}, \bm{\theta}^\mathbf{k}, \bm{\theta}^\mu\}$, where $\bm{\theta}^\mathbf{M}$ represents the mass and inertia, $\bm{\theta}^\mathbf{k}$ represents restitution, and $\bm{\theta}^\mu$ represents friction coefficients.
Under the rigid-body assumption where object motion follows the Newton-Euler equations,  these parameters 
are sufficient to define collision, inertial response, and contact behavior~\cite{featherstone2014rigid, 3DLCP}. Properties like elasticity or plasticity fall outside the rigid-body scope.

The differentiable rendering function $\mathcal{I}$ is used to align the visual observations between the source domain and the target domain. Note that from a differentiable physics perspective, the floor is stationary, and the robot is the one applying force actively, whose dynamics will not be affected by other objects, so only the motion of the target object needs to be observed and aligned. 
Therefore, our rendering function only consider the target object, that is $\mathcal{I}(\mathbf{x}_t, \bm{\alpha}) = \mathcal{I}(\mathbf{x}_t^o, \bm{\alpha})$, where $\bm{\alpha}$ represents the rendering parameters specifically defined for the target object, and the rendered image will change with the update of object pose. 

The learning process of our \method{} starts with optimizing $\bm{\alpha}$ for rendering alignment, and uses optimized $\bm{\alpha}^*$ to guide the identification of physical parameters $\bm{\theta}$ for simulation.

\paragraph*{\textbf{Rendering Alignment}} 
To optimize the rendering parameters $\bm{\alpha}$ for the target object $o$, the robot end-effector moves around $o$ in its initial state $\mathbf{x}_0^o$ and captures multiple static scene images with an eye-in-hand camera $\textbf{I}^s = \{I^s_0, ..., I^s_m\}$, as demonstrated in Figure~\ref{fig:pipeline}(a). 
To make sure that the rendering function $\mathcal{I}$ can generalize to new viewpoints or object poses,
we adopt 2D Gaussian Splatting (2DGS)~\cite{2DGS} as the render.
Compared to 3D Gaussian splatting~\cite{kerbl20233d}, 2DGS is more effective in capturing surface details.

2DGS renders images by optimizing a set of Gaussian elliptical disks,  
which are defined in local tangent $uv$ planes in world space:
\begin{equation}
    P(u, v) = \mathbf{p}_k + s_u \mathbf{t}_u u + s_v \mathbf{t}_v v = \mathbf{H}(u, v, 1, 1)^\top,
\end{equation}
where $\mathbf{p}_k$  is the central point of the $k$-th 2D splat.
$\mathbf{t}_u$ and $\mathbf{t}_v$ are principal tangential vectors, and $\mathbf{t}_w = \mathbf{t}_u \times \mathbf{t}_v$ presents the primitive normal. $\mathbf{R} = [\mathbf{t}_u, \mathbf{t}_v, \mathbf{t}_w]$ is a $3 \times 3$ rotation matrix and \( \mathbf{S} = (s_u, s_v) \) is the scaling vector.
The 2D Gaussian for static object representation 
can be equivalently represented by a homogeneous matrix $\mathbf{H}_0 \in 4\times4$:
\begin{equation}
    \mathbf{H}_0 = \begin{bmatrix}
    s_u\mathbf{t}_u & s_v\mathbf{t}_v & \mathbf{0} & \mathbf{p}_k  \\
     0 & 0 & 0 & 1
    \end{bmatrix} = \begin{bmatrix}
    \mathbf{RS} & \mathbf{p}_k \\
    \mathbf{0} & 1
    \end{bmatrix}.
\end{equation}

During the optimization, we randomly splat  2D disks onto the object surface $G$ for high-quality rendering initialization.  
For an image coordinate $(x, y)$, volumetric alpha blending integrates alpha-weighted appearance to render the image $\hat{I}$:
\begin{equation}\label{eq:appearance}
\hat{I}(x, y) = \sum_{i=1}  \bm{\alpha}^c_i \bm{\alpha}^o_i
\mathcal{G}_i(\mathbf{u}(x, y)) \prod_{j=1}^{i-1} \left(1 - \bm{\alpha}^o_j 
\mathcal{G}_j(\mathbf{u}(x, y))\right),
\end{equation}
where $\bm{\alpha}^c_i$ and $\bm{\alpha}^o_i$ are the color and opacity of the $i$-th Gaussian, 
$\mathbf{u}(x, y)$ represents the intersection point between the ray emitted from the camera viewpoint through the image pixel $(x, y)$ and the plane where the 2D Gaussian distribution resides in 3D space, 
$\mathcal{G}(\mathbf{u})$ is the 2D Gaussian value for intersection $\mathbf{u}$, indicating its weight.

This differentiable rendering models the visual observation of the target object $o$ in its initial state, and the corresponding parameters $\bm{\alpha}$, including all 2DGS parameters, are optimized with the following loss function:
\begin{equation}\label{eq:appearance_loss}
    \mathcal{L} = \mathcal{L}_c + \omega_d \mathcal{L}_d + \omega_n \mathcal{L}_n,
\end{equation}
where $\mathcal{L}_c$ combines rendering loss~\cite{nerf} 
$\mathcal{L}_r = \| \hat{\mathbf{I}} - \mathbf{I}^s \|_2^2$  with the D-SSIM term~\cite{kerbl20233d}. $\mathcal{L}_d$ and $\mathcal{L}_n$ are regularization terms for depth distortion and normal consistency~\cite{2DGS}, respectively.

Once the optimal parameters $\bm{\alpha}^*$ are obtained for the target object $o$ in its initial state  $\mathbf{x}_0^o$, any change of the object pose can lead to the rendering updating, achieved by transforming the 2DGS accordingly. In more detail, for any new object state $\mathbf{x}_t^o$, we can convert its corresponding pose to a transformation matrix $\mathbf{T}_{t}^o\in 4\times4$~\cite{hartley2003multiple}. 
We then apply $\mathbf{T}_{t}^o$ to initial Gaussian splats represented by $\mathbf{H}_0$, resulting in the transformed homogeneous matrix:
 \begin{equation}
    \mathbf{H}_{t} = \mathbf{T}^o_{t}(\mathbf{T}_{0}^o)^{-1}\mathbf{H}_0.
\end{equation}
where $\mathbf{T}^o_{0}$ represents the static object pose in its initial state  $\mathbf{x}_0^o$ as the reference. $\mathbf{H}_{t}$ can be used to render the new image for the target object with an updated state, denoted as $ \mathcal{I}  (\mathbf{x}_{t}, \bm{\alpha}^*)$.

\paragraph*{\textbf{Identification of Physics Parameters}}
With the optimized rendering parameter $\bm{\alpha}^*$, we further estimate the physics properties $\bm{\theta}$ for simulation, based on the gradient flow from visual observations established by the differential render. 
The robot interacts with the object in state $\mathbf{x}_{t}$ through a set of task-agnostic actions $\mathcal{A} =  \{\mathbf{a}_{t}, ..., \mathbf{a}_{t+n-1}\}$ to collect a video $\mathbf{I}^d = \{I^d_{t+1}, ..., I^d_{t+n}\}$ capturing dynamics, as shown in Figure~\ref{fig:pipeline}(b). 
The transformed observations $\hat{\mathbf{I}} = \{\mathcal{I}(\mathbf{x}_{t+i}, \bm{\alpha}^*)\}_{i=1}^{n}$ 
are then obtained in simulation with Equation~\ref{eq:appearance}, where $\mathbf{x}_{t+i} = g(\mathbf{x}_{t+i-1}, \mathbf{a}_{t+i-1}, \bm{\theta})$ is the updated state when applying action $\mathbf{a}_{t+i-1}$. 
The physics parameter $\bm{\theta}$ is then 
estimated by minimizing the discrepancy between the generation $\hat{\mathbf{I}}$ and observation $\mathbf{I}^d$.
Therefore, we can represent the objective of the physics estimation as:
\begin{equation}\label{eq:objective}
    \min_{\bm{\theta}} 
    \quad \mathcal{L}_r(\bm{\theta}) = \sum_{i=1}^{n} \| 
    \mathcal{I}   (g(\mathbf{x}_{t+i-1}, \mathbf{a}_{t+i-1}, \bm{\theta}), \bm{\alpha}^*) - I^d_{t+i} \|_2^2.
\end{equation}

What remains now is to develop a differentiable physics model $\mathbf{x}_{t+1} = g(\mathbf{x}_{t}, \mathbf{a}_t,
\bm{\theta})$  
for simulation,  predicting the next object pose $\mathbf{x}_{t+1}$  based on 
current state $\mathbf{x}_{t}$ and action $\mathbf{a}_t$. 
Note that previous work estimates physics parameters by differentiating the impact of external wrenches on objects~\cite{3DLCP}. 
However, for robotic manipulation, we cannot assume that all robot parts support wrench measuring. Therefore, we choose the 
translation $\mathbf{d}_t$ 
of robot end-effector as the action, i.e., $\mathbf{a}_t = \mathbf{d}_t$.

We formulate this system identification process as a  velocity-based Linear Complementarity Problem (LCP)~\cite{2DLCP, 3DLCP} which solves the equations of motion under global constraints. 
Here, we use LCP to first estimate $\bm{\xi}_{t+1}$ from $\mathbf{x}_t$, and then further use those two together to update the remaining $\mathbf{p}_{t+1}$ and $\mathbf{q}_{t+1}$.
In more detail, given a time horizon $H$ which describes the duration of an action's effect, LCP  updates twist velocities of each scene component $\bm{\xi}_{t}$ 
to $\bm{\xi}_{t+1}$ after $H$, where $\bm{\xi}_{t}$ includes linear velocities $\mathbf{v}_t$ 
and angular velocities $\bm{\Omega}_t$. 
The updated velocity $\bm{\xi}_{t+1} = \{\mathbf{v}_{t+1}, \bm{\Omega}_{t+1}\}$ is then used to calculate the updated pose $\{\mathbf{p}_{t+1}, \mathbf{q}_{t+1}\}$ integrated by the semi-implicit Euler method~\cite{li2023difffr}:
\begin{align}\label{eq:physics}
    \mathbf{p}_{t+1} & = \mathbf{p}_{t} + H \cdot \mathbf{v}_{t+1}, \notag \\ 
    \mathbf{q}_{t+1} & = \text{normalize}(\mathbf{q}_{t} + \frac{H}{2} ([0, \bm{\Omega}_{t+1}]\otimes \mathbf{q}_{t}) ),
\end{align}
where $\mathbf{p}_t$ and $\mathbf{q}_t$ are the object's position and orientation represented by a quaternion.   
$[0, \bm{\Omega}_{n+1}]$ represent quaternion constructed from the angular velocity $\bm{\Omega}_{n+1}$, and $\otimes$ denotes the quaternion multiplication.

The LCP is solved following the framework by~\citet{cline2002rigid},  where the goal is to find velocities $\bm{\xi}_{t+1}$ and Lagrange multipliers $\bm{\lambda}_e, \bm{\lambda}_c, \bm{\lambda}_f, \bm{\gamma}$
satisfying the momentum conservation  when a set of constraints is included:
\begin{align}
    &\bm{\theta}^\mathbf{M} \bm{\xi}_{t+1} = \bm{\theta}^\mathbf{M} \bm{\xi}_t + \mathbf{f}^{\text{g}} \cdot H + \mathbf{J}_e \bm{\lambda}_e + \mathbf{J}_c\bm{\lambda}_c + \mathbf{J}_f\bm{\lambda}_f, \notag 
    \\
    & \,\,\quad\quad\quad\quad\quad\quad\quad\quad\quad\quad\quad \text{(Rigid Body Dynamics Equation)} \notag 
    \\
    & \mathbf{J}_e \bm{\xi}_{t+1} = 0, \quad\quad\quad\quad\quad\quad\quad\quad\quad\quad\,\,\,\quad\quad \text{(Joint Constraints)}  \notag \\
    & \mathbf{J}_c \bm{\xi}_{t+1} \geq - \bm{\theta}^\mathbf{k} \mathbf{J}_c\bm{\xi}_t \geq -\mathbf{c}, \quad\quad\quad\quad\quad\,\,\,   \text{(Contact Constraints)}  \notag \\
    & \mathbf{J}_f \bm{\xi}_{t+1} + \mathbf{E} \bm{\gamma} \geq 0, \quad \bm{\theta}^\mu \bm{\lambda}_c \geq \mathbf{E}^\top \bm{\lambda}_f,  \quad \text{(Friction Constraints)} 
\end{align}
where $\mathbf{f}^\text{g}$ is the gravity wrench, $\bm{\lambda}_e, \bm{\lambda}_c, \bm{\lambda}_f, \bm{\gamma}$ are constraint impulse magnitudes,  $\mathbf{E}$ is a binary matrix making the equation linearly independent at multiple contacts, and $\mathbf{J}_e, \mathbf{J}_c, \mathbf{J}_f$ are input Jacobian matrices describing the joint, contact, and friction constraints, please refer to  \cite{2DLCP} for construction details.
Here, joint constraints ensure that connected objects maintain a specific relative pose, contact constraints prevent interpenetration, and friction constraints enforce the maximum energy dissipation principle.

We adopt the primal-dual interior point method~\cite{mattingley2012cvxgen} as the LCP solver to obtain the solution $\bm{\xi}_{t+1}$ while establishing gradient propagation from $\bm{\xi}_{t+1}$ to $\bm{\theta}$. 
We then  apply the method described in \cite{amos2017optnet} to derive the gradients of the solution 
with the objective in Equation~\ref{eq:objective}.
The output of the physics model $g$ is contributed by both $\bm{\theta}$ and the last state $\mathbf{x}_t$, where $\mathbf{x}_t$ also depends on $\bm{\theta}$.
Therefore, the gradients with respect to $\bm{\theta}$ can be expressed as:
\begin{align}\label{eq:gradient}
    \frac{d\mathcal{L}_r(\bm{\theta})}{d\bm{\theta}} = \sum_{i=1}^{n}   ( 
        \mathcal{I} & 
    (g(\mathbf{x}_{t+i-1},  \mathbf{a}_{t+i-1}, 
         \bm{\theta}), \bm{\alpha}^*)  - I^d_{t+i} ) \notag \\
        & \cdot(\frac{\partial \mathcal{I}}{\partial g} 
        \frac{\partial g} {\partial \bm{\theta}}  +  \frac{\partial \mathcal{I}}{\partial g} \frac{\partial g}{\partial \mathbf{x}_{t+i-1}} \frac{\partial \mathbf{x}_{t+i-1}}{\partial \bm{\theta}}).
\end{align}
As LCP is widely adopted by mainstream simulators~\cite{coumans2016pybullet, MakoviychukWGLS21}, the estimated $\bm{\theta}$ can be compatible with existing simulation environments~\cite{james2020rlbench, chen2022towards} as well.

Since we use a velocity-based LCP,  the end-effector translation $\mathbf{d}$ can be converted into velocity $\bm{\xi}^{e} = \mathbf{d} / H$, where $H$ is the action time horizon.
This equation holds because the robot's mass is typically much greater than the object's mass, allowing its own dynamics to be ignored. The pose and velocity of the floor is kept stationary during the whole process, and only its geometry and physical parameters are used for solving the dynamics equation.
Moreover, in robot manipulation, the action time horizon $H$ is usually inequivalent to simulation step size $h$, while the latter is set small enough to ensure accurate object pose integration (Equation~\ref{eq:physics}). 
The input sub-action for each recursion is derived by dividing the original action $\mathbf{a}_t$ into $H / h$ segments.
We propagate recursive derivatives of Equation~\ref{eq:gradient}  across $H / h$  simulation time steps and optimize $\bm{\theta}$.

\subsection{Physics-aware Digital Cousins}

The learned world model $\mathcal{W}$ reduces the gap with real worlds and provides an interactive environment for manipulation policy learning.
However, inconsistencies with the real world remain due to inaccurate and partial observations.
Domain randomization~\cite{visionDM}  randomizes system parameters in the source domain during training to cover the problem space of the target, but it often lacks adequate constraints, increasing training burdens and reducing policy performance. Therefore, We propose \emph{physics-aware digital cousins}, which perturb the rendering and physics near the system’s identified parameters, as illustrated in Figure~\ref{fig:pipeline}(c).

We adopt all estimated rendering parameters $\bm{\alpha}^*$ and physics parameters $\bm{\theta}^*$ for generating digital cousins.
For rendering, we adopt the spherical harmonics (SH) parameters 
$\bm{\alpha}^{\text{sh}} \subset \bm{\alpha}^*$~\cite{muller2022instant}, which represent the directional rendering component of the 2D Gaussian.
Perturbing $\bm{\alpha}^{\text{sh}}$ allows modeling of lighting and material variations. 
We randomize the system parameters $\bm{\omega}^r = \{\bm{\alpha}^{\text{sh}}, \bm{\theta}^*\}$ by sampling $\tilde{\bm{\omega}}^r$ from a uniform distribution: 
\begin{equation}\label{eq:randomization}
    \tilde{\bm{\omega}}^r \sim \mathcal{U}(\bm{\omega}^r \cdot (1 - \delta), \bm{\omega}^r \cdot ( 1 + \delta)) 
\end{equation}
where $\delta$ indicates perturbation magnitude. For SH parameters, randomization is applied separately to each splat.
To ensure zero-shot policy transfer,
we perturb the identified parameters with $\delta = 0.1$ to generate  digital cousins.
 Our physics-informed world model is compatible with arbitrary reinforcement learning methods; we adopt Proximal Policy Optimization (PPO)~\cite{schulman2017proximal} for its ease of implementation.
 The learned policy is then directly deployed in the target domain world for manipulation tasks, as shown in Figure~\ref{fig:pipeline}(d).

%% file: result.tex
\section{Results and Evaluations}
\label{sec:results}

With our experimental evaluations, we aim to answer the following questions:

\begin{itemize}
    \item Does our method 
     outperform other Real2Sim2Real methods in learning deployable manipulation policies?
      \item Does \method{} achieve more accurate system identification compared to existing approaches?
    \item Does the proposed physics-aware digital cousins (PADC) help with policy transfer?
     \item Can our method deliver superior performance in real-world settings?
\end{itemize}

We conduct experimental evaluations in both simulation and the real world.
Simulators provide ground truth for evaluating system identification accuracy and hence offer comprehensive answers to the first three questions, while the real-world tests are used to validate the effectiveness of policy deployment regarding the last question.

We evaluate our method on rigid body motion control.
The robot's objective is to perform a sequence of non-prehensile actions to move an object into a target pose. Actions are specified as translations of the end-effector.
We set up two tasks: \emph{push}~\cite{chi2023diffusion} and \emph{flip}~\cite{zhou2023hacman}. 
The push task is to move a planar object on a plane to a target pose, involving 2D translation in the $xy$-plane and 1D rotation around the $z$-axis.
The flip task is to poke an object to turn it from a lying pose to an upside-down pose, which requires 3D rotation and 3D translation.

\subsection{Evaluations in Simulation}

\begin{figure}[tp]
    \centering
    \begin{overpic}[width=0.45\textwidth, tics=5]{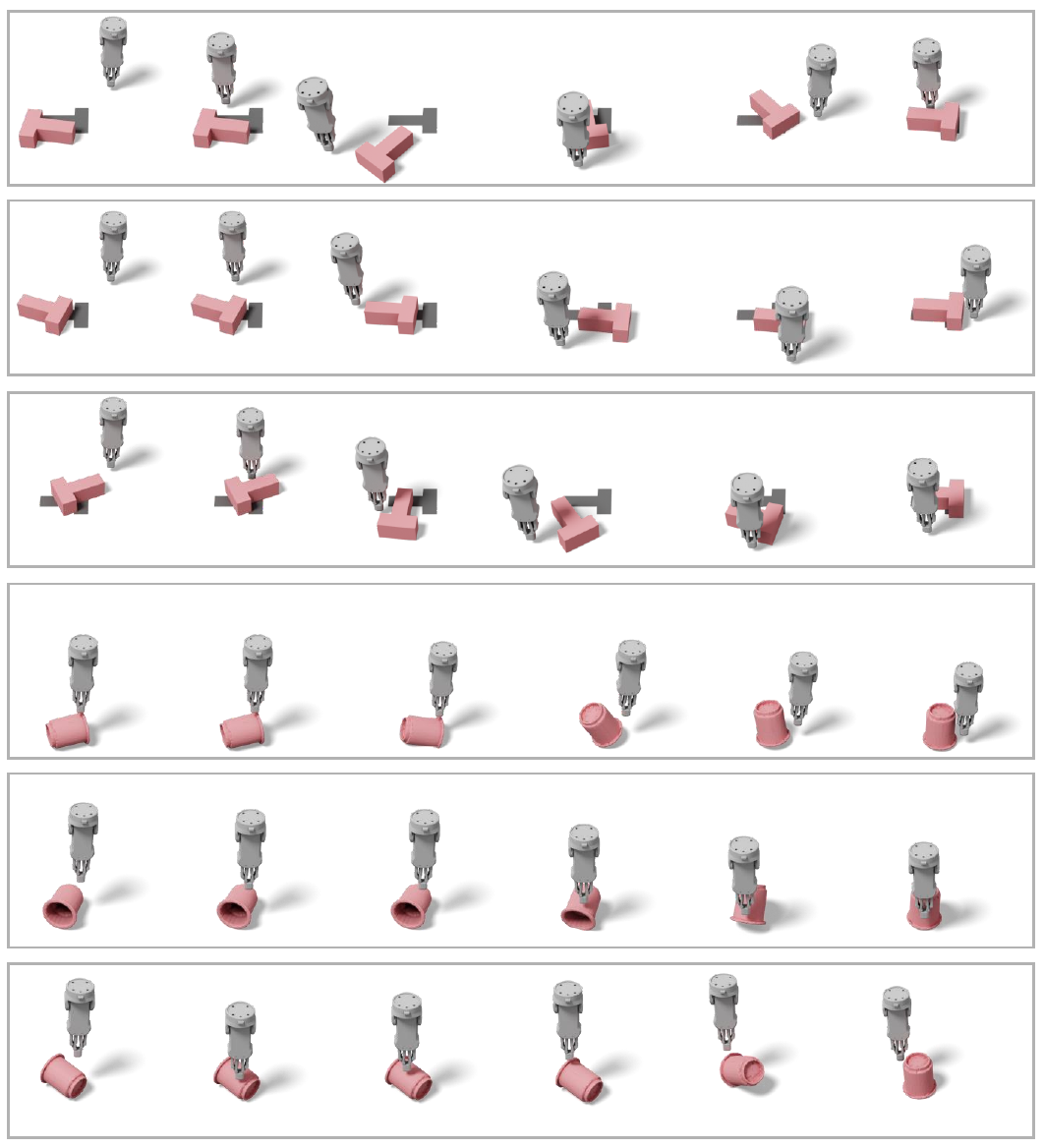}
        \put(-3,75){\rotatebox{90}{\makebox(0,0){\small Push tasks}}}
        \put(-3,25){\rotatebox{90}{\makebox(0,0){\small Flip tasks}}}
        \end{overpic}
    \caption{
        Manipulation trajectories in simulation obtained by our method for both \emph{push} and \emph{flip} tasks.
    }
    \label{fig:sim}
\vspace{-10pt}
\end{figure}

\paragraph*{\textbf{Experiment setup}}
In simulation, we collect a \emph{single} task-agnostic trajectory that the target object is pushed forward along a straight line by the robot end-effector for a predefined distance in the target domain.
After that, any access to the target domain is prohibited.
Since our estimated parameters are compatible with existing simulators, we integrate estimations to the Bullet engine~\cite{coumans2016pybullet} for high-performance physics simulation. 
With the learned simulator, we train manipulation policies with only RGB images as input. 
We maintain 32 parallel threads for efficient training, each running an independent physics-aware digital cousin.
The initial object pose is randomized for each episode.
After the episode terminates for each thread, the environment is replaced with a newly sampled digital cousin.
Trained policies are then directly deployed to the target domain for evaluation. 
For both push and flip tasks, we set a relatively low friction in the target domain to highlight the importance of physics identification.
Figure~\ref{fig:sim} demonstrates several manipulation trajectories obtained by our method for both push and flip tasks with different initial states.

\paragraph*{\textbf{Evaluation metrics}}
To answer the first three questions, our evaluation metrics focus on both the manipulation policies and the world models. 
We measure the success rate $Succ\, \%$ of a policy if the task is completed within a threshold of $100$ steps for push and $25$ steps for flip. 
We also consider the required number of steps to complete a task, denoted as $\#Steps$.
We evaluate the accuracy of a world model using \textit{one-step error}~\cite{lambert2022investigating}
which measures the distance between the final object states after applying one sampled action to the identified model and the target-domain simulator. 
This error is computed separately for translation and rotation differences, measured in meters and radians, respectively.

\paragraph*{\textbf{Baseline methods}}
We compare our method with various types of approaches for training non-prehensile manipulation skills, 
including: \\
\indent $\bullet$  \textit{Methods that rely purely on data.} A representative is the well-known Dreamer V2~\cite{HafnerL0B21}, which is a latent-space dynamics model from data for handling high-dimensional observations and learning robust policies. Given their strong reliance on data quantity, we provide $100$ task-agnostic trajectories. Based on the learned expert policy, 
we train non-prehensile manipulation skills via imitating expert demonstrations~\cite{hussein2017imitation} of $100$ \emph{task-completion} trajectories similar to~\citet{chi2023diffusion}. \\
\indent $\bullet$  \textit{Methods with pre-defined physics-based world models.} 
We use Bullet~\cite{coumans2016pybullet} as the default simulator. Following the standard domain randomization approach~\cite{dynamicsDM}, we randomize the physics parameters in Bullet, including mass, friction, restitution, and inertia, across a broad range for policy training. We employ our learned rendering function $\mathcal{I}$ as the renderer.
We set a variant with fixed, random physics and rendering parameters where no system identification or randomization is involved, denoted as \emph{Random}. We also compare with RoboGSim~\cite{li2024robogsim}, which optimizes only rendering parameters using 3DGS~\cite{kerbl20233d} but not physics parameters. \\
\indent $\bullet$  \textit{Methods with learned physics-based world models.}  We compare with ASID~\cite{memmel2024asid} and the method of \citet{SongB20}. The former performs system identification using gradient-free optimization. The latter leverages differentiable 2D physics (thus referred to as 2D Physics).
Since neither of the two methods learns rendering parameters and their trained policies cannot work without aligned visual input, we add our rendering function $\mathcal{I}$ to enhance these two methods.\\
\indent Note that all physics-based methods being compared are trained with the same task-agnostic trajectories as \method, for fair comparison. 
All policies are trained until no significant success rate performance can be gained and are then deployed directly to the target domain for evaluation.
We also conduct an ablation study of our method that trains policies without PADC.
More implementation details of baseline methods are provided in Appendix~\ref{sec:baselines}.

\input{tables/policy_transfer.tex}

\paragraph*{\textbf{Comparisons on policy performance}}

We conduct $100$ episodes of tests for each method and report the comparison results in Table~\ref{tab:policy_results}.
Our method achieves the best performance for both non-prehensile manipulation tasks, thanks to the accurate system identification of PIN-WM and the meaningful digital cousins of PADC.
Without PADC, our method still outperforms others, although with a performance decrease.

The purely data-driven world model Dreamer V2~\cite{HafnerL0B21}, albeit having access to more task-agnostic data, fails to accurately approximate the dynamics of the target domain, resulting in poor performance of the trained and deployed policies.
Diffusion Policy~\cite{chi2023diffusion}, relies on more expensive task-completion data, also presents inferior performance due to the limited training data quantity and hence poor out-of-distribution generalization.
These results highlight the importance of incorporating physics priors in learning world models.

For those methods with pre-defined physics-based world model, Domain Rand~\cite{dynamicsDM} + $\mathcal{I}$ introduces excessive randomness, making the task learning more difficult. 
We provide a detailed explanation in Appendix~\ref{sec:more_results}  of why Domain Rand  + $\mathcal{I}$ struggles; smaller-scale randomizations around ground-truth physical parameters improve its effectiveness, though knowing these parameters is unrealistic.
RoboGSim~\cite{li2024robogsim} optimizes only for rendering parameters but not physics ones, also leading to performance degradation. 
In contrast, our physics-aware digital cousin design, perturbing the physics and rendering parameters around the identified values as means, creates meaningful digital cousins allowing for learning robust policies with Sim2Real transferability.

Moreover, the policies trained with physics-based alternatives exhibit unsatisfactory performance in the target domain.
One reason is that their world models failed to effectively capture the target-domain dynamics.
ASID~\cite{memmel2024asid} leads to suboptimal solutions due to its inefficient gradient-free optimization.
Although 2D Physics~\cite{SongB20} accounts for 2D differentiable physics and achieves satisfactory push performance, its performance on the \emph{flip} task degrades since it involves 3D rigid body dynamics.
These experiments collectively demonstrate that an accurate identification of both physics and rendering parameters is crucial for learning non-prehensile manipulation skills.
Although \method{} performs physical parameter identification under the assumption of perfect geometry, we also provide experiments in Appendix~\ref{sec:more_results} showing that, even with noise, the training environment constructed by PIN-WM effectively supports policy training.

\paragraph*{\textbf{Comparisons on system identification}}
We compare the accuracy of system identification of both data-driven and physics-based approaches. This is done by measuring the one-step error after applying the same randomly sampled action to the same surface point 
of the target object. The results are reported in Table~\ref{tab:world_model_accuracy}.

\input{tables/model_accuracy_origin.tex}

\begin{figure}[t]
    \begin{minipage}{0.5\textwidth}
    \centering
    % \hfill
    \subfigure{
        \includegraphics[width=0.47\linewidth]{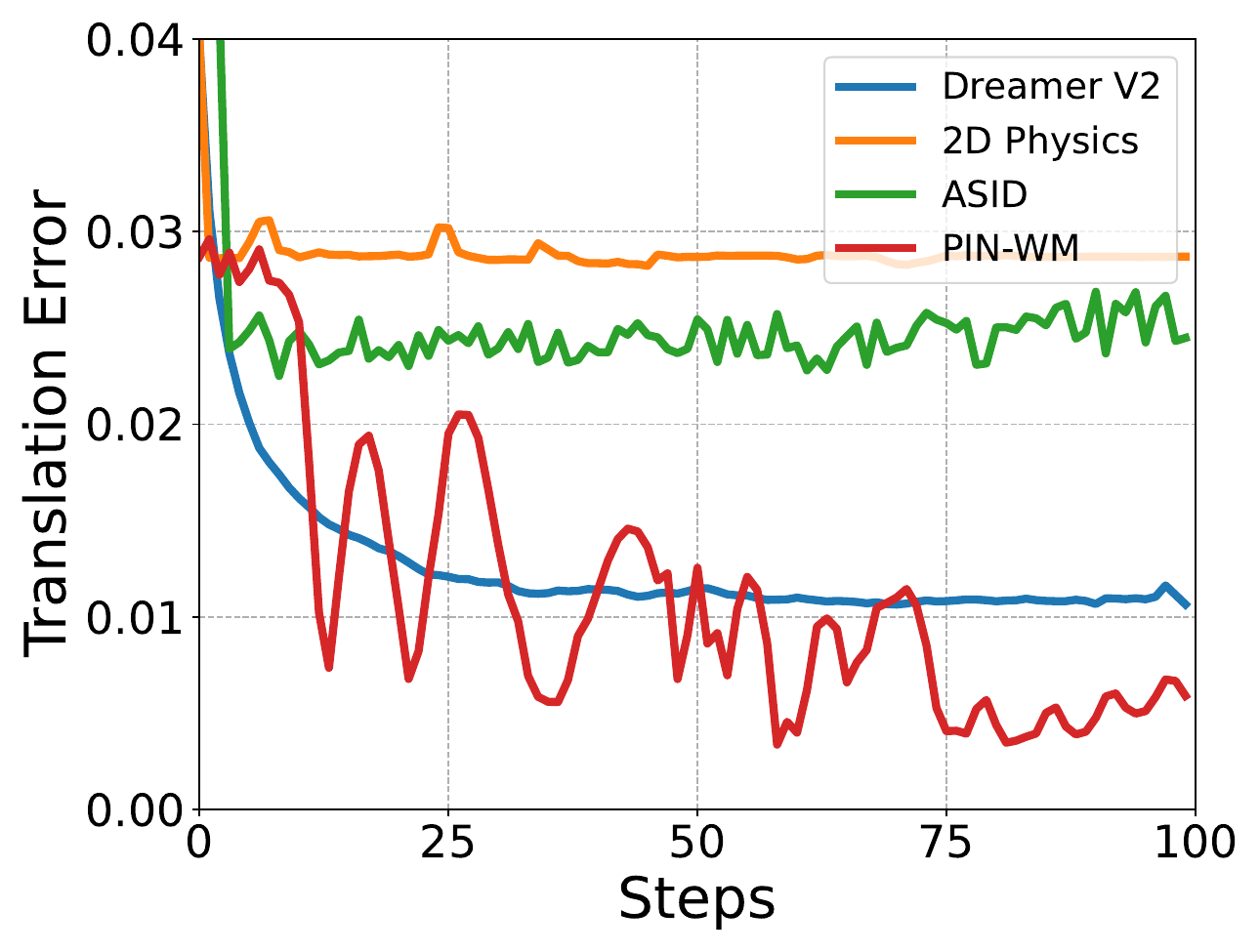}}
\subfigure{
    \includegraphics[width=0.47\linewidth]{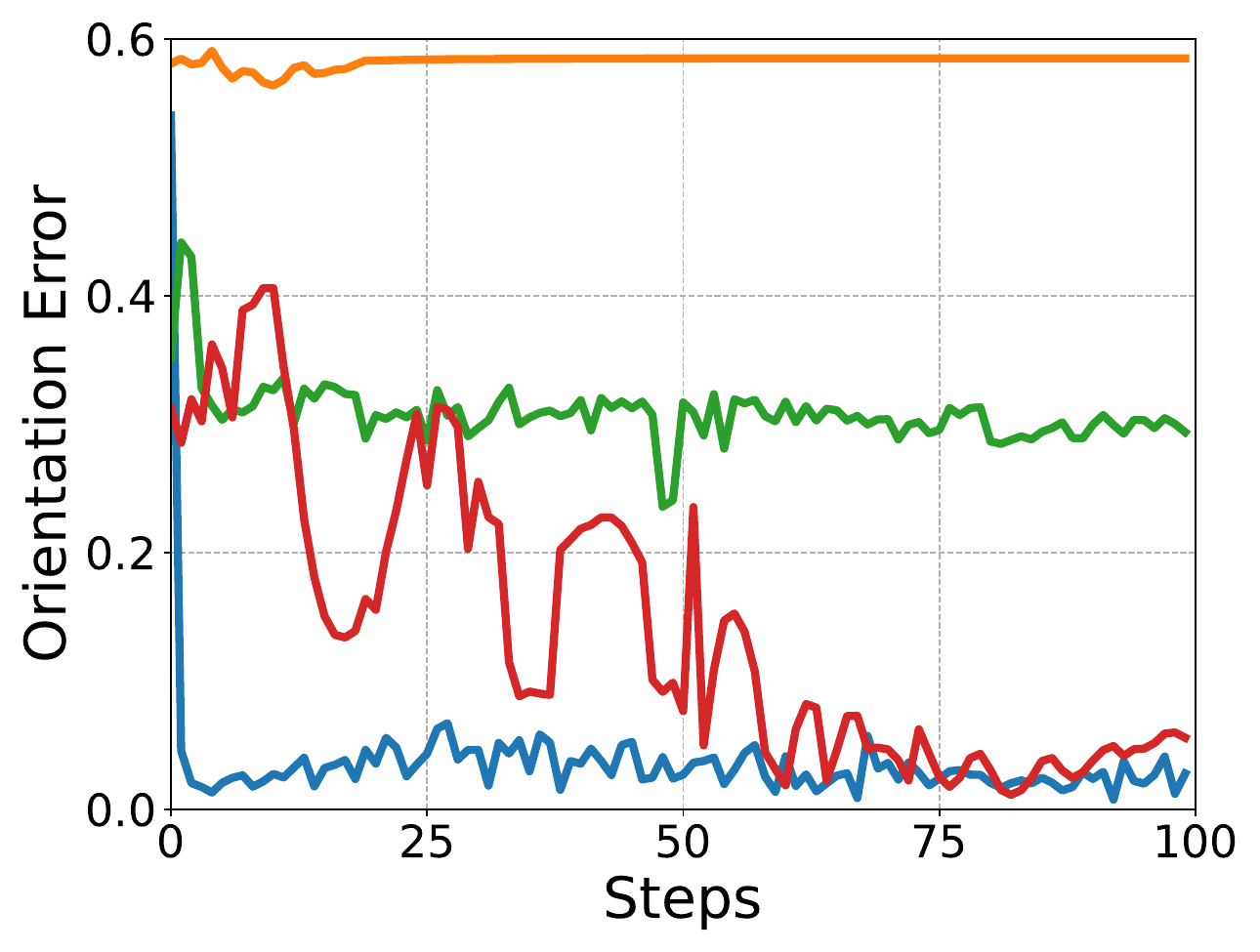}}
    \vspace{-10pt}
    \caption{
        Transition and orientation errors of  \emph{push} task during training.
        % of the learned dynamics model.
        }
    \label{fig:world_model_learning}
    \end{minipage}
\vspace{-10pt}
\end{figure}

We observe that the data-driven method Dreamer V2~\cite{HafnerL0B21}, as expected,  
suffer catastrophic performance degradation when generalizing to new state and action distributions. 
ASID~\cite{memmel2024asid} shows lower accuracy compared to \method{} in both push and flip tasks, since it is difficult for gradient-free optimization to find a good solution in a finite time due to the large search space.
Although 2D Physics~\cite{SongB20} adopts a differentiable framework, the 2D model finds difficulties in handling 3D rigid body dynamics, resulting in low prediction accuracy.
In contrast, our method learns 3D rigid body physics parameters through differentiable optimization, achieving superior performance in both push and flip scenarios.
We present the learning curves of the push task in Figure \ref{fig:world_model_learning}, demonstrating the stability and efficiency of \method{} during training. 
We can observe that Dreamer V2 quickly converges on the training dataset, but it does not generalize well on the test dataset.
We also provide the physical parameters identified by each method, along with the ground truth parameters, in Appendix~\ref{sec:more_results}.

\subsection{Evaluations in Real-World}
\label{sec:real_world_experiment}
\begin{figure}[h!t]
    \centering
    \includegraphics[width=\linewidth]{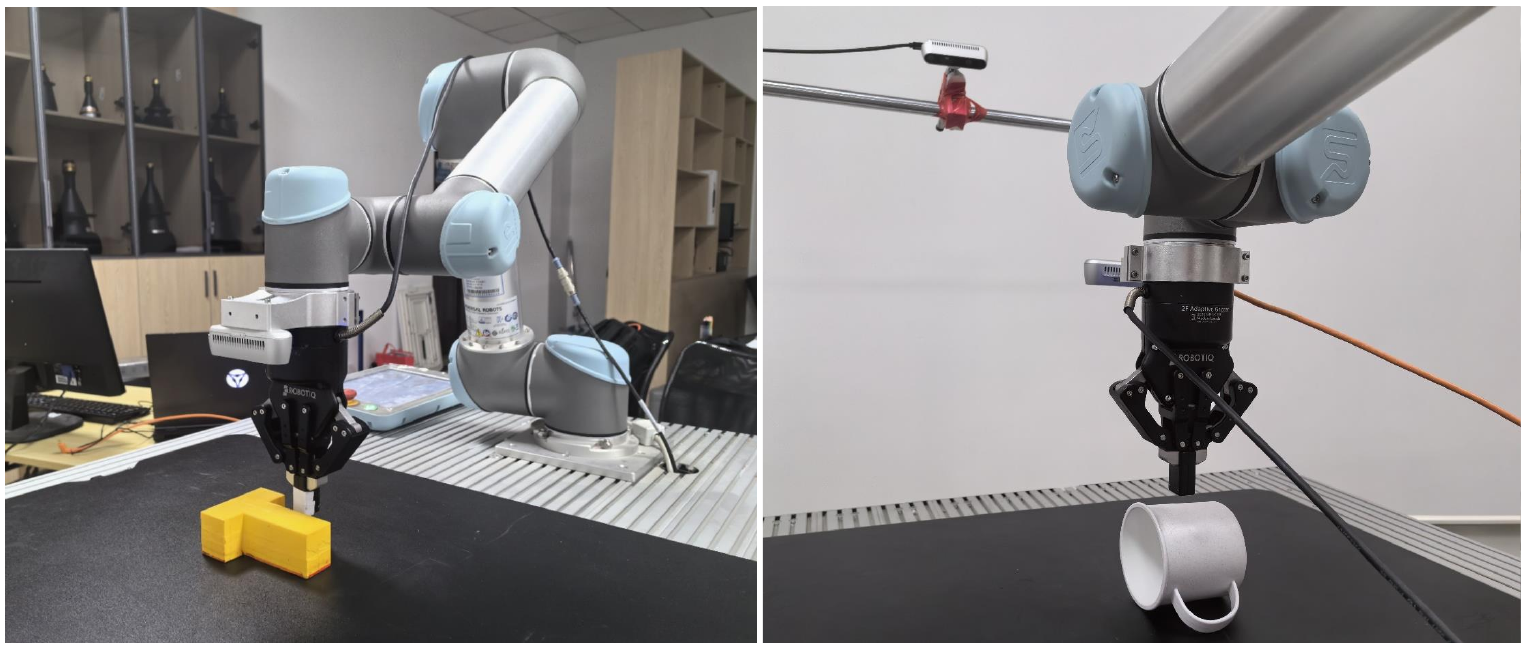}
    \caption{
        Our real-world experiment setup.
    }
    \label{fig:real_robot}
\end{figure}

\paragraph*{\textbf{Experiment setup}}
Our hardware setup consists of a robot, an eye-in-hand camera, and an eye-to-hand camera, as shown in Figure~\ref{fig:real_robot}. Given a real-world object $o$ and its 
mesh geometry $G$, we use FoundationPose~\cite{wen2024foundationpose} 
to estimate initial object pose $\mathbf{T}^o_0$ in the world coordinate system. We set the mesh geometry $G$ with pose $\mathbf{T}^o_0$ in the simulator, and sample a series of surface points on the transformed mesh $\mathbf{T}^o_tG$ as the initialization for 2D Gaussian Splatting. The robot then moves around the object and captures the time-lapse video sequence $\mathbf{I}^s$, while recording the corresponding camera pose sequence $\{ \mathbf{T}_0^c, ..., \mathbf{T}_n^c \}$. We segment the region of interest with SAM 2~\cite{ravi2024sam} and optimize the 2D Gaussian with the objective in Equation~\ref{eq:appearance_loss}, aligning rendering with the real world.
After that, we apply 
a straight line of translational actions to push the object $o$ in the real world.
A dynamic video $\mathbf{I}^d$ is captured by the eye-to-hand camera
to be used for optimizing the physics parameters $\bm{\theta}$ with Equation~\ref{eq:objective}.

\input{tables/real_world.tex}

\paragraph*{\textbf{Baseline methods}}
In the real-world setting, collecting a large amount of trjactories, either task-agnostic or task-completion, is highly expensive. Therefore, we only compare with policy learning methods requiring \emph{no or few-shot real-world data}, thus excluding data-driven methods such as Dreamer V2~\cite{HafnerL0B21} and Diffusion Policy~\cite{chi2023diffusion}.

\begin{figure*}[!t]
    \centering % 添加居中命令
    % \hspace{0.5cm}
    \begin{overpic}[width=0.95\textwidth,tics=5]{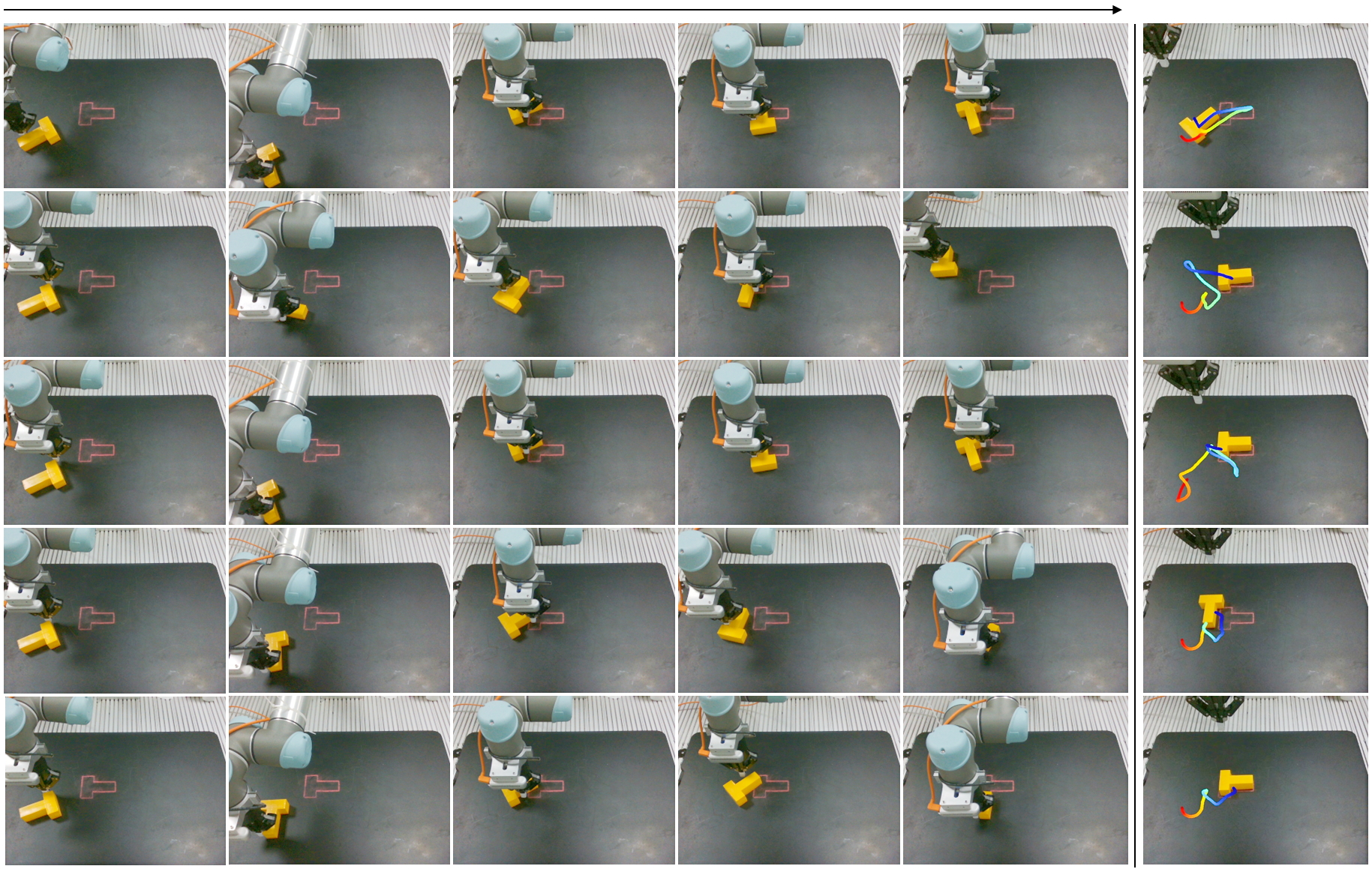}
    \put(-3,55){\rotatebox{90}{\makebox(0,0){\small Domain Rand.}}}
    \put(-3,42){\rotatebox{90}{\makebox(0,0){\small RoboGSim}}}
    \put(-3,31){\rotatebox{90}{\makebox(0,0){\small ASID }}}
    \put(-3,18){\rotatebox{90}{\makebox(0,0){\small 2D Physics}}}
    \put(-3,6){\rotatebox{90}{\makebox(0,0){\small PIN-WM (ours)}}}
    \put(40,63.5){\rotatebox{0}{\makebox(0,0){ Time lapse}}}
    \put(90,63.5){\rotatebox{0}{\makebox(0,0){ Result}}}
    \end{overpic}
    \caption{
    Real-world trajectories of different methods on the \emph{push} task. 
    }
    \label{fig:real-world-traj}
\end{figure*}

\begin{figure*}[!t]
\centering % 添加居中命令
\vspace{10pt}
\begin{overpic}[width=0.95\textwidth,tics=5]{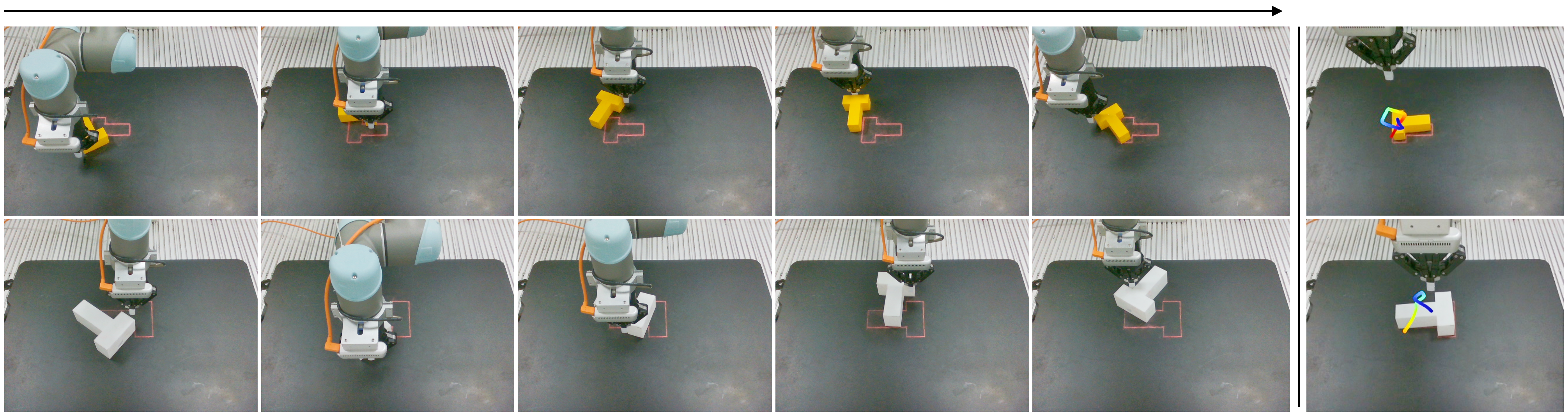}
\put(-3,5){\rotatebox{90}{\makebox(0,0){\small Large T}}}
\put(-3,18){\rotatebox{90}{\makebox(0,0){\small Small T}}}
\put(40,27){\rotatebox{0}{\makebox(0,0){ Time lapse }}}
\put(90,27){\rotatebox{0}{\makebox(0,0){ Result}}}
\end{overpic}
\caption{
Real-world trajectories of pushing T-shaped objects of different sizes obtained by our method. 
}
% \vspace{-10pt}
\label{fig:diff-t}
\end{figure*}
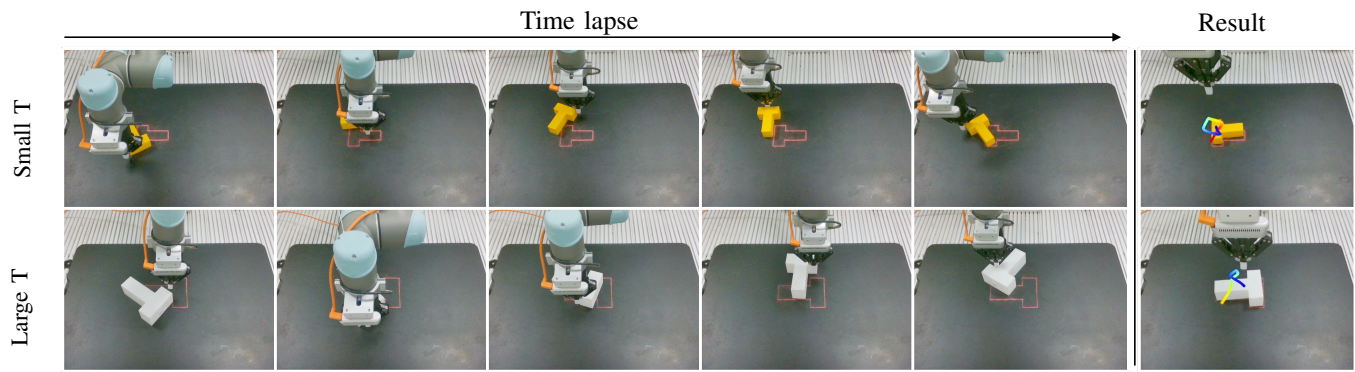

\paragraph*{\textbf{Comparisons on policy performance}}
We evaluate real-world performance with both \emph{push} and \emph{flip} tasks under identical initial conditions across 20 trials.
The push task requires pushing the T-shaped object to the red target position, while the flip task involves flipping a mug from its side to an upside-down pose.
The results are summarized in Table~\ref{tab:real_world}, showing that \method{} can complete the task with higher success rates and fewer steps.

Conventional simulators with random parameters fail to produce transferable policies due to physical and rendering misalignment. 
Although having integrated our rendering function $\mathcal{I}$, the naive randomization of Domain Rand~\cite{dynamicsDM} still creates noisy variations that degrade policy learning.
This is also demonstrated by the comparisons with RoboGSim~\cite{dynamicsDM}, which aligns rendering but not physics parameters.
With a more accurate estimation of physics parameters, 2D Physics~\cite{SongB20} and ASID~\cite{memmel2024asid} obtain slightly better results but are still inferior to \method{}.
By learning both physical parameters and rendering representations through differentiable optimization,
together with our physics-aware digital cousins design, our approach attains much better performance in real-world deployments.

We show comparisons of real-world trajectories of \emph{push} task in Figure~\ref{fig:real-world-traj}.
Our method successfully pushes the T-shaped object to the target pose with a few steps. In contrast, alternative approaches either require longer trajectories or fail to complete the task.
We also verify the effectiveness of our method by demonstrating how it completes the \emph{push} task on a larger T-shaped object in Figure~\ref{fig:diff-t},  demonstrating its adaptability to varied shapes and sizes.
Appendix~\ref{sec:further_real_world} provides real-world comparisons for pushing objects on a slippery glass plane, including both the T-shaped object and a cube object.
We provide trajectories about flipping a mug in Figure~\ref{fig:real-world-flip} and a cube object in Appendix~\ref{sec:further_real_world}.

\begin{figure}[h]
    \centering
    \begin{overpic}[width=0.46\textwidth,tics=5]{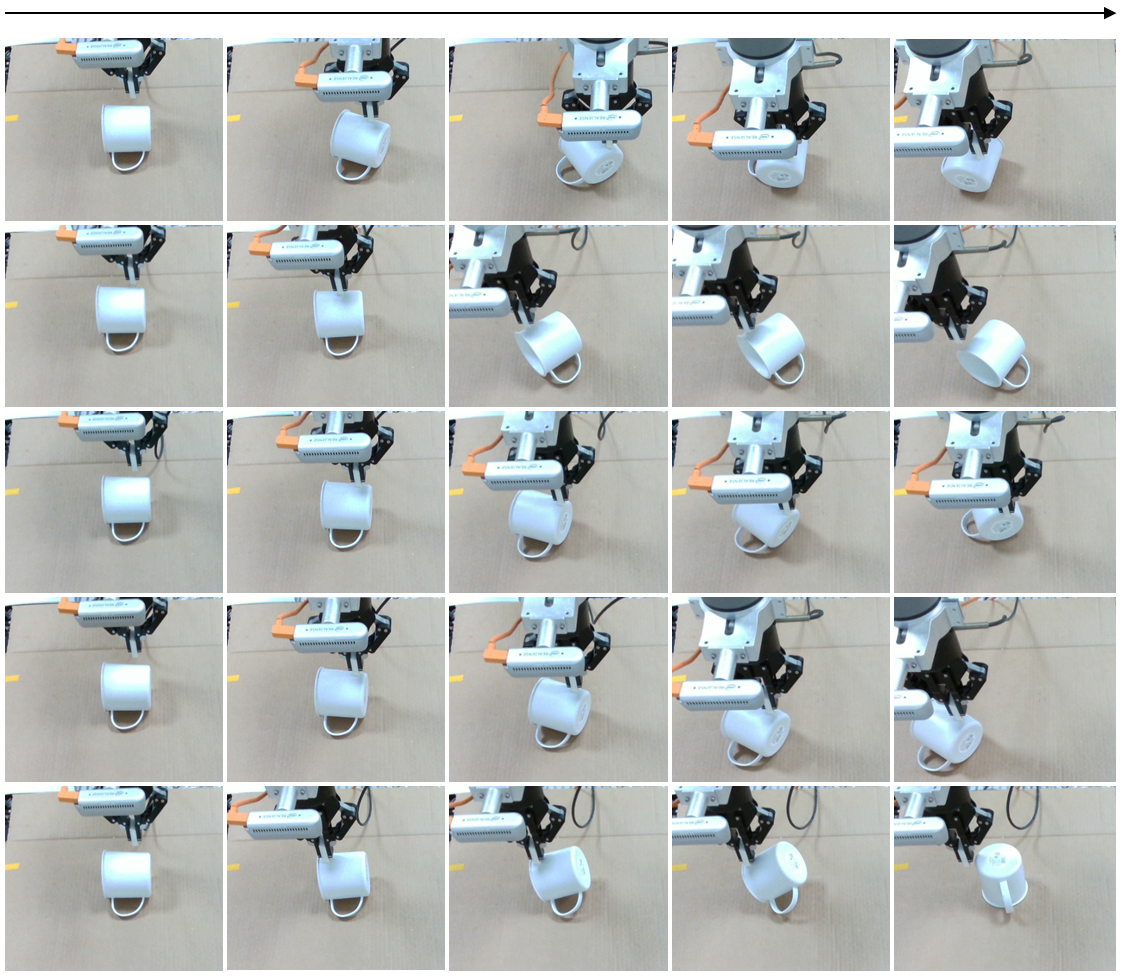}
    \put(-3,77){\rotatebox{90}{\makebox(0,0){\footnotesize Domain Rand.}}}
    \put(-3,57){\rotatebox{90}{\makebox(0,0){\footnotesize RoboGSim}}}
    \put(-3,42){\rotatebox{90}{\makebox(0,0){\footnotesize ASID }}}
    \put(-3,25){\rotatebox{90}{\makebox(0,0){\footnotesize 2D Physics}}}
    \put(-3,8){\rotatebox{90}{\makebox(0,0){\footnotesize PIN-WM }}}
    \put(50,88){\rotatebox{0}{\makebox(0,0){ Time lapse}}}

    \end{overpic}
    \caption{Real-world trajectories of different methods on the \emph{flip} task.}
    \label{fig:real-world-flip}
  \end{figure}

%% file: tables/policy_transfer.tex
\begin{table}[t]
    \centering
    \caption{
        Comparisons on policy performance in the target domain.
        }
    \label{tab:policy_results}
    \resizebox{0.5\textwidth}{!}{
    \begin{tabular}{c|cc|cc}
    \toprule
    \multirow{3}*{\textbf{Methods}} & \multicolumn{4}{c}
    {\textbf{Tasks}}
    \\
    &\multicolumn{2}{c|}{\textit{Push}}& \multicolumn{2}{c}{\textit{Flip}} 
    \\
    & \multicolumn{1}{c}{$Succ\, \%$} & \multicolumn{1}{c}{$\#Steps$} &  \multicolumn{1}{|c}{$Succ\, \%$} & \multicolumn{1}{c}{$\#Steps$} 
    \\
    \midrule
    \multirow{1}*{ Random
    }
    &  0\% & 100.0 & 0\% & 25.0
    \\
    \multirow{1}*{Dreamer V2~\cite{HafnerL0B21}}  & 1\% & 99.9 & 0\% & 25.0
    \\
    \multirow{1}*{Diffusion Policy~\cite{chi2023diffusion}}  & 13\% & 91.1 & 10\% & 23.3 
    \\
    \multirow{1}*{RoboGSim~\cite{li2024robogsim}}  & 19\% & 82.6 & 
    21\% & 20.5
    \\
    \multirow{1}*{Domain Rand~\cite{dynamicsDM} + $\mathcal{I}$ }  & 33\% & 73.6 & 
    32\% & 20.0
    \\
    \multirow{1}*{2D Physics~\cite{SongB20}} + $\mathcal{I}$  & 55\% & 60.6 &8\% & 22.6
    \\
    \multirow{1}*{ASID~\cite{memmel2024asid} + $\mathcal{I}$}  & 58\% & 57.6 & 
    11\% & 22.2
    \\
    \midrule
    \multirow{1}*{\method{} w/o PADC}  & 92\% & 32.1 & 70\% & 12.0
    \\
    \multirow{1}*{\method{} w/ PADC} & $\mathbf{97}\%$ & $\mathbf{30.1}$ & $\mathbf{83}\%$ & $\mathbf{11.4}$
    \\
    \bottomrule
\end{tabular}
}
% \vspace{-10pt}
\end{table}

%% file: tables/model_accuracy_origin.tex
\begin{table}[t]
    \centering
    \caption{
       Comparisons on system identification accuracy across different methods,
          using 
         \emph{one-step error} of the predicted trajectory. ``Trans.'' and ``Rot.'' are translation and rotation errors, respectively. 
        } 
    \label{tab:world_model_accuracy}
    \resizebox{0.5\textwidth}{!}{
    \setlength{\tabcolsep}{2mm}{
    \begin{tabular}{c|cc|cc
        }
    \toprule
    \multirow{3}*{\textbf{Methods}} & \multicolumn{4}{c}{\textbf{Tasks}}\\
     & \multicolumn{2}{c|}{\textit{Push}} & \multicolumn{2}{c}{\textit{Flip}} 
     \\
     & \multicolumn{1}{c}{Trans.} &  \multicolumn{1}{c|}{Ori.} & \multicolumn{1}{c}{Trans.} &  \multicolumn{1}{c}{Ori.}
    \\
    \midrule
    \multirow{1}*{Dreamer V2~\cite{HafnerL0B21}} 
    & 7.6$\times 10^{-2}$ 
    & 5.6$\times 10^{-1}$ & 2.3$\times 10^{-1}$ & 2.0 
    \\
    \multirow{1}*{2D Physics~\cite{SongB20}}  
    & 4.2$\times 10^{-2}$  
    & 5.4$\times 10^{-1}$  & 2.1$\times 10^{-1}$ & 1.6 
    \\
    \multirow{1}*{ASID~\cite{memmel2024asid}}  
    & 3.0$\times 10^{-2}$ 
    & 4.0$\times 10^{-1}$ & 1.4$\times 10^{-1}$ & 1.6 
    \\
    \multirow{1}*{\method}  
    & $\mathbf{1.7\times 10^{-2}}$ 
    & $\mathbf{1.3\times 10^{-1}}$  & $\mathbf{1.4\times 10^{-2}}$ & $\mathbf{0.3}$
    \\
    \bottomrule
\end{tabular}

}}
\end{table}

%% file: tables/real_world.tex
\begin{table}[t]
    \centering
    \caption{
    Real-world deployment performance.}
    \begin{tabular}{c|cc|cc}
    \toprule
    \multirow{3}*{\textbf{Methods}} 
    & \multicolumn{4}{c}{\textbf{Tasks}}\\
    &\multicolumn{2}{c|}{\textit{Push}} & \multicolumn{2}{c}{\textit{Flip}} 
    \\
    & $Succ \%$ & $\#Steps$ & $Succ \%$ & $\#Steps$    \\
    \midrule
    \multirow{1}*{Random} & 0\% & 100.0 & 0\% & 25.0
    \\
    \multirow{1}*{Domain Rand~\cite{dynamicsDM} + $\mathcal{I}$} 
    & 10\% & 87.5 & 25\% & 18.5
    \\
    \multirow{1}*{RoboGSim~\cite{li2024robogsim}} 
    &30\%  & 72.7 & 15\% & 21.2
    \\
    \multirow{1}*{2D Physics~\cite{SongB20} + $\mathcal{I}$} 
    &35\%  & 70.7 & 5\% & 24.1
    \\
        \multirow{1}*{ASID~\cite{memmel2024asid} + $\mathcal{I}$}   &40\%  & 64.6 & 10\% & 22.4 
    \\

    \midrule
    \multirow{1}*{\method{} w/o PADC}  & 65\%  & 45.2 & 60\% & 13.2 
    \\
    \multirow{1}*{\method{} w/ PADC}  & \textbf{75}\% & \textbf{37.5} & \textbf{65}\% & \textbf{11.3}
    \\
    \bottomrule
\end{tabular}
\label{tab:real_world}
\end{table}

%% file: conclusion.tex
\section{Conclusions} 
\label{sec:conclusion}

We have presented a method of end-to-end learning physics-informed world models of 3D rigid body dynamics from visual observations. Our method is able to identify the 3D physics parameters critical to non-prehensile manipulations with few-shot and task-agnostic interaction trajectories. To realize Sim2Real transfer, we turn the identified digital twin to a group of physics-aware digital cousins through perturbing the physics and rendering parameters around the identified mean values. Experiments demonstrate the robustness and effectiveness of our method when compared with different types of baseline methods.

\paragraph*{\textbf{Limitation and future work}}
We see several opportunities for future research.
First, we use visual observations to guide the optimization of physical parameters, thus rendering alignment places a key role here. We find that the various shadows generated with the robot's movement can distort rendering loss estimation, compromising the accuracy of learned physical properties. 
This issue could potentially be resolved by incorporating differentiable relighting~\cite{gao2024relightable} into Gaussian Splatting to better model lighting conditions.
Additionally, our current framework focus on rigid-body dynamics, and it would be interesting to explore ways to integrate more advanced differentiable physics engines, such as the Material Point Method (MPM)~\cite{MurthyMGVPWCPXE21}, to extend \method{}'s capability to handle deformable objects.  
We are also engaged in applying PIN-WM to real-world applications in industrial automation~\cite{TangA0WHWFS0N24, zhao2023learning,  zhao2025deliberate}.

\section{Acknowledgements}
This work was supported in part by the NSFC (62325211, 62132021, 62322207), the Major Program of Xiangjiang Laboratory (23XJ01009),  Key R\&D Program of Wuhan  (2024060702030143), Shenzhen University Natural Sciences 2035 Program (2022C007), and Guangdong Laboratory of ArtificialIntelligence and Digital Economy Open Research Fund (GML-KF-24-35).

%% file: appendix.tex
\appendix

\subsection{Implementation  Details for Baselines}
\label{sec:baselines}
All the baselines are implemented carefully to ensure fair comparison. 
We used the official implementations with default hyperparameters for Diffusion Policy~\cite{chi2023diffusion}, 2D Physics~\cite{SongB20}, and Dreamer V2~\cite{HafnerL0B21}. A history of recent states and actions is used as input for the ``Domain Rand + $\mathcal{I}$" (denoted as DR) baseline~\cite{dynamicsDM}.
All RL-based policies are trained using PPO~\cite{schulman2017proximal}, with the same model architecture, reward function,  hyperparameters, and stopping criterion based on the success rate.
The reward signal for policy learning is a handcrafted function to encourage the robot to push the object toward the target pose: $r = -d_t-d_r$, where $d_t$ and $d_r$ refer to the translation distance and rotation distance, respectively.
Diffusion Policy is trained with successful trajectories collected from expert policies trained in the environment with GT physical parameters, without any randomization.

\subsection{More Experimental Results}
\label{sec:more_results}
\subsubsection{{Superiority over uniform randomization}}
The only difference between our method and the DR baseline is the different ranges of physics parameters that are used for domain randomization, where DR uses a large range $\mathbf{R}$ to ensure that it covers the target parameters $\bm{\theta}^\dag$ while our method uses a much smaller range around the learned parameters $\bm{\theta}^*$. The low performance of DR is mainly due to the large range of $\mathbf{R}$ and the performance can be improved by shrinking the range around $\bm{\theta}^\dag$ (even though this is not feasible in real application scenarios), and using GT $\bm{\theta}^\dag$ directly can obtain the best result, as presented in Table~\ref{tab:randomization_range}.
\input{tables/success_rate.tex}

\subsubsection{{
  Identified physical parameters}} 
System identification is inherently ill-posed, as multiple parameter sets can explain the same observations. To fairly compare the accuracy across different methods, we estimate one parameter at a time while keeping the others fixed at their GT values. 
Since inertia is typically represented as a $3\times3$ matrix in simulation, we do not conduct this experiment.
Our method consistently achieves the best performance, as shown in Table~\ref{tab:world_model_parameters}.
\input{tables/model_accuracy.tex}

\subsubsection{{Robustness to geometry noise}} 
Geometry noise will affect collision detection accuracy and, consequently, system identification.
To evaluate this, we conduct experiments on noisy inputs, in which the noise conforms to a Gaussian distribution with a mean of 0 and a variance of $\sigma$\%$L$, where $\sigma \in \{0, 0.5, 1.0, 3.0\}$ and  $L$ is the length of object's bounding box diagonal.
We report one-step error of the predicted trajectories after applying the same random actions to the object. As shown in Table~\ref{tab:noise}, while the prediction errors increase with higher noise, our method remains robust even at $\sigma=3.0$, outperforming ASID~\cite{memmel2024asid} using perfect geometry.
\input{tables/noise.tex}

\subsection{Further Real-World Evaluations}
\label{sec:further_real_world}

\begin{figure}[h]
  \centering
  \begin{overpic}[width=0.38\textwidth,tics=5]{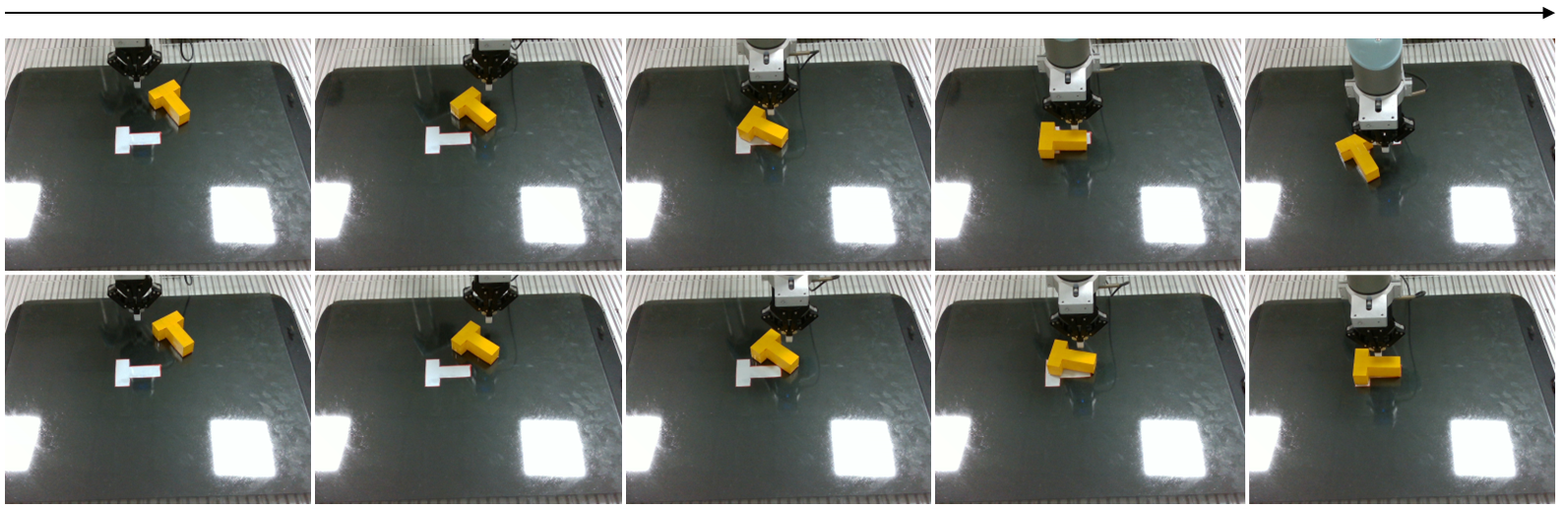}
  \put(50,34){\rotatebox{0}{\makebox(0,0){ Time lapse}}}
  \put(-3,22.5){\rotatebox{90}{\makebox(0,0){\footnotesize ASID }}}
  \put(-3,7.5){\rotatebox{90}{\makebox(0,0){\footnotesize PIN-WM }}}
  \end{overpic}
  \caption{\emph{Push} T-shaped object on a slippery plane. }
  \label{fig:slippery_t}
  \vspace{-10pt}
\end{figure}

\begin{figure}[h]
  \centering
  \begin{overpic}[width=0.38\textwidth,tics=5]{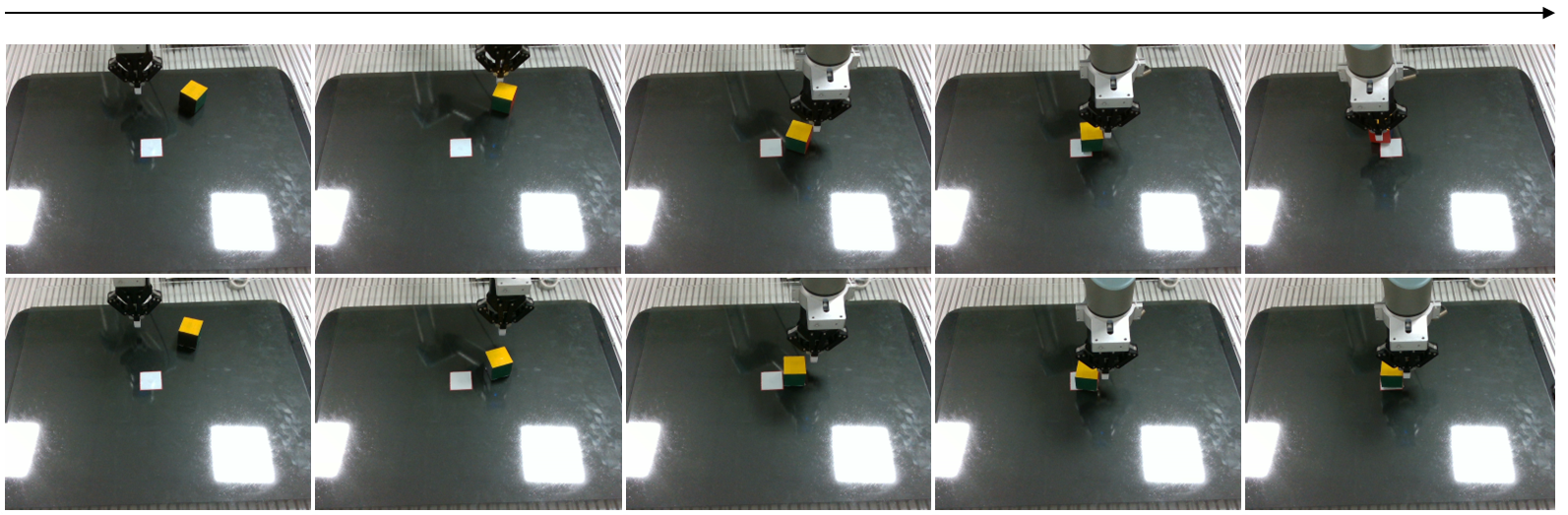}
    \put(50,35){\rotatebox{0}{\makebox(0,0){ Time lapse}}}
    \put(-3,22.5){\rotatebox{90}{\makebox(0,0){\footnotesize ASID }}}
    \put(-3,7.5){\rotatebox{90}{\makebox(0,0){\footnotesize PIN-WM }}}
  \end{overpic}
  \caption{\emph{Push} cube object on a slippery plane. }
  \label{fig:slippery_square}
  \vspace{-10pt}
\end{figure}

\begin{figure}[h!]
  \centering
  \begin{overpic}[width=0.38\textwidth,tics=5]{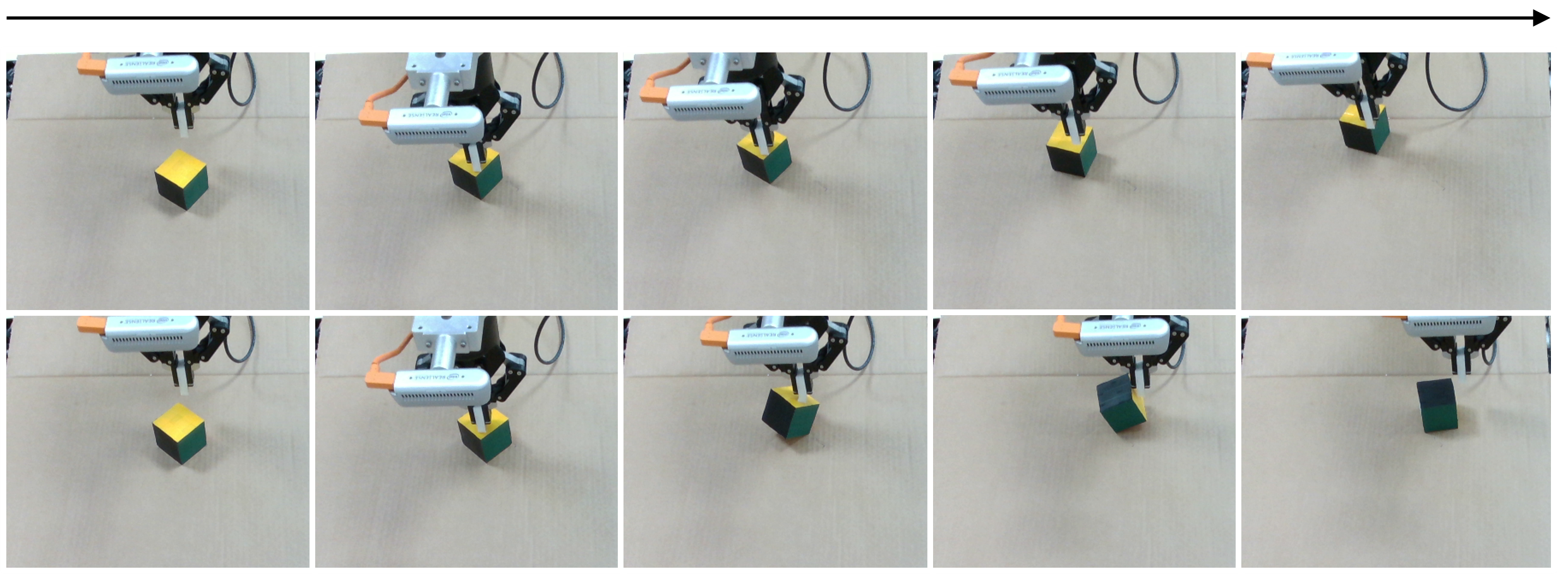}
        \put(50,38){\rotatebox{0}{\makebox(0,0){ Time lapse}}}
    \put(-3,24.5){\rotatebox{90}{\makebox(0,0){\footnotesize ASID }}}
    \put(-3,8.5){\rotatebox{90}{\makebox(0,0){\footnotesize PIN-WM }}}
  \end{overpic}
  \caption{\emph{Flip} a multicolored cube to change its top-surface color. }
  \label{fig:real-world-flip2}
  \vspace{-10pt}
\end{figure}

We further validate the effectiveness of \method{} in identifying real-world physical parameters. We push a T-shaped object and a cube object on a slippery glass plane. Even small touches cause noticeable displacement, which poses higher demands on the robot’s control precision. The smaller mass and volume of the cube make it harder to manipulate.
 Real-world trajectories are  provided in Figures~\ref{fig:slippery_t} and~\ref{fig:slippery_square}, separately. PIN-WM demonstrates strong performance on both objects, successfully pushing them to the target positions. In contrast, the baseline ASID consistently pushes the objects with excessive force just before reaching the target position, making it difficult to complete the task. We also conduct flip experiments on the small cube, with the trajectories provided in Figure~\ref{fig:real-world-flip2}. The quantitative results of these experiments are summarized in Table~\ref{tab:more_tasks}.

 \input{tables/tasks.tex}

%% file: tables/success_rate.tex
\begin{table}[h]
    \centering
    \vspace{-10pt}
    \caption{Success rates across different ranges of randomization.}
    \vspace{-4pt}
    \resizebox{0.5\textwidth}{!}{
    \setlength{\tabcolsep}{1mm}{
    \begin{tabular}{c|cccccccc}
    \toprule
    \multirow{2}*{} 
    \textbf{Tasks} & GT & PIN-WM w/ PADC & DR ($\mathbf{R}$/4) &  DR ($\mathbf{R}$/2)  &   DR ($\mathbf{R}$)  &
    \\
    \midrule
    \multirow{1}*{\textit{Push}}  
    & $\mathbf{98\%}$
    & 97\% & 78\% & 56\% &  33\%
    \\
    \multirow{1}*{\textit{Flip}}  
    &  $\mathbf{89\%}$    &  83\%  &  61\%   &  43\%  &   32\%
    \\
    \bottomrule
\end{tabular}
}
}
\vspace{-10pt}
\label{tab:randomization_range}
\end{table}

%% file: tables/model_accuracy.tex
\begin{table}[h]
    \centering
    \vspace{-10pt}
    \caption{Identified physical parameters.}
    \vspace{-4pt}
    \resizebox{0.5\textwidth}{!}{
    \setlength{\tabcolsep}{0.5mm}{
    \begin{tabular}{c|ccc|ccccc}
    \toprule
    \multirow{2}*{\textbf{Methods}} 
     & \multicolumn{3}{c|}{\textit{Push}} & \multicolumn{3}{c}{\textit{Flip}} 
    \\
    & Friction & Mass (kg) & Restitution 
    & Friction & Mass (kg) & Restitution 
    \\
    \midrule
    GT & $3.00 \times 10^{-2}$ & 1.00 & 0.00  & $2.00 \times 10^{-1}$ & 1.00 &  0.00 
    \\
    \midrule
    \multirow{1}*{ASID}  
    & {$4.45 \times 10^{-2}$} 
    & {0.75} & $1.12 \times 10^{-2}$  
    & $2.59 \times 10^{-1}$ & 1.36 &   $1.12  \times 10^{-2}$ 
    \\
    \multirow{1}*{2D Physics}  
    & {--}  
    & 19.25 & {--}  
    % & {--} 
    & {--}  
    & 29.60 & {--}  
    % & {--}
    \\
    \multirow{1}*{PIN-WM}  
    &$\mathbf{3.05 \times 10^{-2}}$ 
    & $\mathbf{0.76}$   & $\mathbf{4.18 \times 10^{-4}}$  & $\mathbf{2.11 \times 10^{-1}}$ & $\mathbf{1.19}$ &  $\mathbf{1.69  \times 10^{-5}}$ 
    \\
    \bottomrule
\end{tabular}
}
}
\vspace{-10pt}
\label{tab:world_model_parameters}
\end{table}

%% file: tables/noise.tex
\begin{table}[h]
    \centering
    \vspace{-10pt}
    \caption{One-step errors across different noise levels.
    }
    \vspace{-4pt}
    \resizebox{0.5\textwidth}{!}{
    \setlength{\tabcolsep}{2mm}{
    \begin{tabular}{c|cc|cc}
    \toprule
    \multirow{2}*{\textbf{Methods}} 
     & \multicolumn{2}{c|}{\textit{Push}} & 
     \multicolumn{2}{c}{\textit{Flip}} 
    \\ & Trans. & Ori. & Trans. & Ori.
    \\ 
    \midrule
    \multirow{1}*{PIN-WM} ($\sigma = 0.0$) 
    & $\mathbf{1.7\times10^{-2}}$ &  $\mathbf{1.3\times10^{-1}}$ & $\mathbf{1.4\times10^{-2}}$ & $\mathbf{2.7\times10^{-1}}$
    
    \\ 
    \multirow{1}*{PIN-WM} ($\sigma = 0.5$) 
    &  $2.1\times10^{-2}$ &  $2.1\times10^{-1}$  & $4.2\times10^{-2}$ & $9.6\times10^{-1}$
    \\ 
    \multirow{1}*{PIN-WM} ($\sigma = 1.0$) 
    &  $2.3\times10^{-2}$ & $1.6\times10^{-1}$ & $4.2\times10^{-2}$ & $9.5\times10^{-1}$
    \\ 
    \multirow{1}*{PIN-WM} ($\sigma = 3.0$) 
    & $2.9\times10^{-2}$ & $3.0\times10^{-1}$ & $7.7\times10^{-2}$ &  $1.5$
    \\ 
    ASID & $3.0\times10^{-2}$ & $4.0\times10^{-1}$ 
    & $1.4\times10^{-1}$ & 1.6
    \\ 
    \bottomrule
\end{tabular}
}
}
\label{tab:noise}
\vspace{-14pt}
\end{table}

%% file: tables/tasks.tex
\begin{table}[h]
    \centering
    \vspace{-4pt}
    \caption{Success rate comparisons on different real-world tasks.}
    \vspace{-4pt}
    \resizebox{0.5\textwidth}{!}{
    \setlength{\tabcolsep}{4mm}{
    \begin{tabular}{c|ccc}
    \toprule
    \multirow{2}*{} 
    \textbf{Methods}  & \multicolumn{1}{c}{Push T (Slippery)}  & Push Cube (Slippery) &  Flip  Cube
    \\
    \midrule
    ASID  
    & $5\%$ &  $0\%$ &  $5\%$
    \\
    \method{} 
    &  $\mathbf{45\%}$    &  $\mathbf{40\%}$ &  $\mathbf{60\%}$  
    \\
    \bottomrule
\end{tabular}
}
}
\vspace{-10pt}
\label{tab:more_tasks}
\end{table}